\documentclass[12pt]{article}
\usepackage[top=1in,bottom=1in,left=1in,right=1in]{geometry}
\usepackage[T1]{fontenc}

\usepackage[
  backend=biber,
  style=authoryear,
  sorting=nyt,
  sortcites=true,
  maxcitenames=2,
  uniquename=false,
  natbib=true          
]{biblatex}
\usepackage[bottom]{footmisc}

\usepackage{caption}
\usepackage[font=footnotesize]{subcaption}

\usepackage{palatino}
\usepackage{amsmath, amsfonts, amsthm}
\usepackage{graphicx}
\usepackage{graphbox}
\usepackage[update,prepend]{epstopdf}
\usepackage{wrapfig}
\usepackage{float}
\usepackage{rotating}
\usepackage{pdfpages}
\usepackage{booktabs}
\usepackage{longtable}
\usepackage{multirow}
\usepackage{multicol}
\usepackage{array}
\usepackage{tabularx}
\usepackage{colortbl}
\newcolumntype{L}[1]{>{\raggedright\arraybackslash}p{#1}}
\usepackage{xcolor}
\definecolor{myLightGray}{RGB}{191,191,191}
\definecolor{myGray}{RGB}{160,160,160}
\definecolor{myDarkGray}{RGB}{144,144,144}
\definecolor{myDarkRed}{RGB}{167,114,115}
\definecolor{myRed}{RGB}{255,58,70}
\definecolor{myGreen}{RGB}{0,255,71}
\definecolor{headerblue}{RGB}{43, 87, 151}
\definecolor{lightgray}{RGB}{220, 230, 241}
\definecolor{inputblue}{RGB}{70,130,180}
\definecolor{featgreen}{RGB}{76,153,100}
\definecolor{axisyellow}{RGB}{210,170,60}
\definecolor{threshred}{RGB}{190,80,80}
\definecolor{trainpurple}{RGB}{120,80,160}
\definecolor{outputgray}{RGB}{90,90,90}
\definecolor{lightbg}{RGB}{245,245,250}
\definecolor{annotgray}{RGB}{100,100,100}
\definecolor{darkgr}{RGB}{60,60,60}
\definecolor{medgr}{RGB}{100,100,100}
\definecolor{litegr}{RGB}{180,180,180}
\definecolor{skipgr}{RGB}{120,120,120}
\usepackage{setspace}
\usepackage{pdflscape}
\usepackage{fancyhdr}
\usepackage[colorlinks, citecolor=black, urlcolor=magenta, linkcolor=black]{hyperref} 
\usepackage{footmisc}
\usepackage{chngcntr}

\usepackage{tikz}
\usetikzlibrary{positioning, arrows.meta, calc, fit, backgrounds, shapes.geometric, decorations.pathreplacing, calligraphy, arrows}
\tikzstyle{bag} = [align=center]
\tikzset{
  state/.style={
    rectangle,
    rounded corners,
    draw=black, very thick,
    minimum height=2em,
    inner sep=2pt,
    text centered,
  },
}
\newcommand{\cmark}[1]{%
  \raisebox{-1pt}{\tikz[baseline=(c.base)]{%
    \node[circle, draw=medgr, fill=medgr!15, inner sep=1.5pt, 
          font=\scriptsize\bfseries, text=darkgr] (c) {#1};}}}

\title{Unpacking the Eye of the Beholder: Social Location, Identity, and the Moving Target of Political Perspectives}

\begin{document}

\title{Unpacking the Eye of the Beholder: \\ Social Location, Identity, and the Moving Target of Political Perspectives\thanks{I am grateful to Andreu Casas, Nora Webb Williams, Alex Todorov, and Donghyeon Won for sharing their datasets with me. All errors are my own.}}

\author{Elena Sirotkina \\
Center for Data Science \\
New York University \\
\href{mailto:es7093@nyu.edu}{es7093@nyu.edu}}

\date{May 2026}

\maketitle

\begin{abstract}
\singlespacing

\noindent Political  and social identities structure how people evaluate political information, a finding decades deep in political science and routinely discarded by computational tools that often produce single scores that treat a piece of text, an image, or a video as if it means the same thing to everyone.  This paper shows that it does not, and that the difference is consequential. To address this problem, I develop the Perspectivist Visual Political Sentiment (PVPS) classifier, which learns from approximately 82,000 evaluations by 5,575 U.S. adults to predict how audiences defined by political and social identities will evaluate the same image. Unlike standard tools that average systematic disagreement away, PVPS preserves it, returning an evaluative profile that records who agrees, who diverges, and along which identity lines. Applied to several influential studies of visual sentiment, PVPS shows that perceived violence in protest imagery and the emotional mechanisms behind protest image engagement both change substantively once audience identity is taken into account. It follows that what a political image conveys is a moving target, and measuring it requires knowing whom it is moving.

\end{abstract}

\newpage
\doublespacing

\section*{Introduction}

The rise of machine learning has given political science a new empirical frontier. Text, images, audio, and video, the media through which political life is increasingly conducted, can now be classified and analyzed at scale, largely drawing on tools from computer vision and NLP that were simply out of reach a decade ago (\cite{grimmer_text_2013, joo_image_2018, Barbera2015Tweeting, lucas2015computer}).

But adoption comes with a price. These tools were built for tasks where a correct answer exists independently of the observer, so the entire infrastructure assumes that disagreement is error to be eliminated. Applied to evaluative tasks - and political sentiment is evaluative at its core - the model learns to predict whichever label the annotator pool produced, performing well by every standard metric, while what was averaged away was the phenomenon political science actually cares about. These are the lines of tension and evaluative conflict that define political disagreement in the first place.

The Perspectivist Visual Political Sentiment (PVPS) classifier developed here is built around this disagreement. Instead of predicting a single sentiment score, it produces an evaluative profile for each image that records which audiences rate it more favorably, along which social and political fault lines, and with what confidence. The classifier is trained on approximately 82,000 ratings of 1,264 political images from 5,575 U.S. adults with full demographic batteries (four survey waves, 2022--2025) and validated against a separate corpus of 7,543 images annotated by over 2,000 workers, drawing on a combined pool of roughly 8,800 unique images and 7,700 respondents.

From this, the classifier learns which visual features of an image predict evaluative divergence between groups, and it applies that learning to images it has never seen. This architecture answers the question that political science actually asks about political images, that where a person stands, politically and socially, determines what they see in a political visual (see e.g. \cite{webb_williams_images_2020, torres_framing_2021}), and the classifier makes that structured variation measurable and predictable at scale.



I use PVPS for Casas and Webb Williams's (\citeyear{casas_images_2019}) finding that enthusiasm mobilizes sharing of Black Lives Matter images. The emotional mechanisms hold, but they operate differently depending on which partisan-demographic group the image favors. On the Democratic Women vs.\ Republican Men axis, enthusiasm produces a steeper mobilization slope for Democrat-women-favorable images, while fear and disgust generate sharply higher engagement for Republican-men-favorable ones. Images that received higher perceived violence scores in Won, Steinert-Threlkeld, and Joo's (\citeyear{won2017protest}) annotation scheme are the same images that the classifier predicts as more favorably evaluated by Republican, conservative, and Republican-leaning respondents.

The contribution of this study runs in both directions across the disciplinary boundary. For political science, the discipline has long established that political evaluation is group-mediated \autocite{campbell_american_1960}, identity-sorted \autocite{mason_uncivil_2018}, and intersectional rather than additive \autocite{crenshaw_mapping_1991, hancock_when_2007}, and yet the empirical tools available have not kept pace with the theory. Surveys aggregate evaluation across groups but human annotation collapses disagreement into a majority label, which looses group-specific predicted reactions across multiple social tension dimensions simultaneously. Here I propose an approach that makes group-mediated evaluation measurable at scale and applicable to corpora that survey methods reach more rarely. 

For computational political science, the field has documented the construct-validity risks that arise when tools built for objective tasks are applied to evaluative ones \autocite{grimmer_textasdata_2022, bestvater_sentiment_2023, gasparyan_decoding_2024}. PVPS shows that the axes of group conflict over political images carry a signal strong enough to be modeled directly from visual content. Once this signal is incorporated into the analysis, the interpretation of how images are received changes substantially. This makes it possible to reanalyze media and visual persuasion effects at scale and re-examine their long-term influence across millions of observations.

\section*{Disagreeing Perspectives in Politics}

Politics is essentially about disagreement. Classic work in political science and sociology treats political conflict as the mobilization of social difference (\cite{schmitt_concept_1932, schattschneider_semisovereign_1960, mouffe_on_2005}), which have crystallized into political divisions that 
outlast the conflicts producing them 
\autocite{lipset_cleavage_1967, rokkan_state_1999, 
bartolini_identity_1990, hooghe_does_2002, kriesi_west_2008}.

These structural divisions are lived through social location, conceived as the core of a person's existence in the social and political world (\cite{Mannheim1936, BergerLuckmann1966}), which underlies \textit{perspectival} disagreement as an unavoidable condition of human social interaction (\cite{kinder_us_2009, mason_uncivil_2018}). As people learn early which social and political groups they belong to (\cite{campbell_american_1960, converse_nature_1964}), these attachments operate as perceptual filters throughout life (\cite{campbell_american_1960, jennings_generations_1981}). The tighter a person's group attachment, the more coherent their political orientations tend to be (\cite{tajfel_integrative_1979}). 

Political science has devoted extensive attention to how social location—e.g., race (\cite{jardina_white_2019}) and ethnicity (\cite{kinder_us_2009}), gender (\cite{phoenix_anger_2019, dugger_social_1988}), class (\cite{wright_class_1997}), and their intersections (\cite{crenshaw_mapping_1991}) - shapes political judgment through group-based heuristics and intuition (\cite{mason_uncivil_2018, haidt_righteous_2012, achen_democracy_2016}). Family, education, and discussion networks reinforce this coherence (\cite{achen_democracy_2016, huckfeldt_citizens_1995}) by strengthening the link between group identity and  judgment. In other words,  where you stand determines what you see, and what you see determines how you judge (\cite{kahan_polarizing_2012, campbell_american_1960, kinder_us_2009}).

Understanding political disagreement, in this tradition, requires understanding the social structure that produces it. But this also implies that if political evaluation is structured by identity and social location, then any attempt to measure it inherits this structure. This is where the rise of computational social science creates both opportunity and risk. 

Computational tools are increasingly used to measure and predict political attitudes\footnote{Or any attitudinal outcome that approximates political attitudes, including sentiment, perceptions, and related evaluative responses.} at scale and these are tasks that depend on human-generated labels and therefore on the same dynamics political science has long studied.  Yet these tools have largely developed without integrating what political science and sociology  already know about the \textit{anticipated} lines of social and political disagreement. The following sections examine how social location and predictive computation intersect, what current approaches miss, and how a perspectivist framework advanced here can address the gap.

\section*{The Moving Target of 'Ground Truth' in Computation}

\subsection*{Making of the "Ground Truth"}

Computational approaches to text, image, and multimodal classification have long operated under a paradigm of ground truth absolutism, wherein the goal of annotation is to converge on a single correct label that represents the "true" state of each data instance (\cite{uma_learning_2021, cabitza_perspectivist_2023, frenda_perspectivist_2025}).

This assumption is built into supervised machine learning itself. Classifiers learn to map inputs to outputs, which means they typically need a single correct answer for each example to measure how well they are doing (\cite{uma_learning_2021}). But when multiple people label the same piece of text, they can disagree. Standard training procedures cannot handle multiple answers at once,\footnote{A model learns by guessing an answer and measuring how wrong it was. 
The standard formula (cross-entropy with a one-hot target) expects 
one correct answer per example, but replacing the one-hot vector 
with the empirical distribution of annotator responses is a 
one-line change that existing frameworks support natively 
\citep{peterson_human_2019}. Majority voting became the default 
not because alternatives are technically infeasible but because 
annotation platforms were built around single-label output and 
most pipelines inherited that convention 
\citep{cabitza_perspectivist_2023}.} so majority voting became the default solution, picking whichever label got the most votes (\cite{cabitza_perspectivist_2023, mostafazadeh_davani_dealing_2022}). The resulting infrastructure was designed to collect redundant labels only to collapse them and treating disagreement as "noise" (\cite{aroyo_truth_2015, dumitrache_crowdsourcing_2018}).

In other words, this "gold standard" framing carried implicit epistemic commitments that categories are ontologically discrete, that competent annotators should agree, and that residual disagreement indicates either task ambiguity requiring clearer guidelines or annotator unreliability requiring filtering (\cite{russell_labelme_2008, raykar_learning_2010}). 

But as more recent and subsequent work would reveal, this convenience came at the cost of erasing exactly the human variation that matters most for subjective tasks (see e.g. \cite{pavlick_inherent_2019, fornaciari_beyond_2021, rottger_two_2022}).

AI fairness, for example, has demonstrated that training data 
is never neutral because when it reflects one group's judgments 
disproportionately, the resulting model reproduces those 
judgments as default (\cite{mehrabi_survey_2021, 
barocas_big_2016}). Applied to annotation, this means that 
a single-label approach does not approximate a universal 
ground truth but privileges whoever is most represented in 
the annotator pool (\cite{prabhakaran_releasing_2021, 
plank_problem_2022}).

Together, these findings raise fundamental concerns about the validity of using aggregated labels as the default approach to training predictive models. When evaluation is shaped by social position, aggregation does not approximate truth but it privileges whoever is most represented in the data.

\subsection*{Whose Ground is More Truth? Perspectivism as a Framework for Computational Political Science}

At the same time, the boundary between tasks where a single ground truth is appropriate and tasks where it is not remains poorly defined. Whenever a task engages political or social identity, annotators bring 
perspectives shaped by who they are. Yet disagreement surfaces even on 
supposedly objective tasks like part-of-speech tagging 
(\cite{plank_problem_2022}) or medical diagnosis 
(\cite{cabitza_perspectivist_2023}), suggesting that ground truth is less 
stable than assumed. Table~\ref{tab:disagreement-taxonomy} summarizes 
the five most frequently cited sources of such disagreement.

In political science, trained coders exhibit systematic disagreement that 
propagates into position estimates (\cite{mikhaylov_coder_2012, 
benoit_crowdsourced_2016}). Partisans assign divergent labels to identical 
images (\cite{williams2026conservatives, gasparyan_media_2025}), 
annotator characteristics shape hate speech judgments 
(\cite{waseem_are_2016, sap_risk_2019}), toxicity ratings 
(\cite{prabhakaran_releasing_2021, sap_annotators_2022, 
goyal_toxicity_2022}), and much of the subjective content  broadly defined
(\cite{mostafazadeh_davani_dealing_2022}).

A growing body of work treats this disagreement as signal rather than 
noise, a position formalized as \textbf{perspectivism} (\cite{cabitza_perspectivist_2023}). 
Rather than collapsing annotations into a single label, perspectivism 
retains who rated each item and what attributes they carry. 
Table~\ref{tab:weak-strong} compares the two levels 
\citet{cabitza_perspectivist_2023} distinguish. Weak perspectivism acknowledges that annotators differ and may weight their contributions unequally, but still produces a single aggregated label. Strong perspectivism abandons aggregation and preserves the full distribution of annotations and treats it as the object of interest.

Yet existing perspectivist work\footnote{Many works do not appear under the perspectivist label, but they effectively implement the same idea.} largely reduces disagreement to a single demographic or social dimension, despite perspectivism being fundamentally concerned with heterogeneous and intersecting evaluations. For example, prior work on textual analysis conditions only on race in toxicity detection (\cite{sap_risk_2019, sap_annotators_2022}) or  demographic groups in hate speech (\cite{fleisig_when_2023}).
In other words,  most perspectivist datasets record attributes in isolation rather than as overlapping identity structures (\cite{orlikowski_ecological_2023}). This is also limiting for political judgment, which emerges at the intersection of political and social identity. 

\begin{landscape}
\begin{table}
\centering
\scriptsize
\captionof{table}{Taxonomy of Annotator Disagreement\textsuperscript{$\dagger$}}
\label{tab:disagreement-taxonomy}
\vspace{0.2cm}
\begin{tabular*}{\linewidth}{@{\extracolsep{\fill}} p{0.2cm} p{4cm} p{10cm} p{8cm}}
\toprule
& \textbf{Source} & \textbf{Characteristics} & \textbf{Resolution} \\
\midrule
\multicolumn{4}{l}{\textit{Resolvable disagreement (reducible through improved design)}} \\[4pt]
1. & Category overlap \newline {\scriptsize (\cite{aroyo_truth_2015})} 
  & Overlapping or ambiguous categories force artificial discretization onto continuous reality, producing systematic label confusion
  & Improve taxonomy and annotation guidelines \\[4pt]
2. & Stimulus quality \newline {\scriptsize (\cite{schaekermann_resolvable_2018})} 
  & Degraded, incomplete, or insufficient stimuli prevent annotators from forming a confident judgment 
  & Improve data quality; filter ambiguous stimuli \\[4pt]
3. & Task design \newline {\scriptsize (\cite{xu_beyond_2025})} 
  & Guidelines, interface, and response format constrain what annotators can express 
  & Redesign annotation protocol \\[4pt]
4. & Reliability variation \newline {\scriptsize (\cite{raykar_learning_2010})} 
  & Annotators differ in competence on tasks with a verifiable correct answer (e.g., named entity recognition, bill passage) 
  & Weight by estimated accuracy; aggregate to single label \\
\midrule
\multicolumn{4}{l}{\textit{Irreducible disagreement (structured by annotator identity)}} \\[4pt]
5. & Identity-structured evaluation \newline {\scriptsize (\cite{sap_risk_2019, sap_annotators_2022, prabhakaran_releasing_2021, gasparyan_media_2025})} 
  & Disagreement is systematic by social location (partisanship, race, gender); reflects genuine evaluative differences rather than error
  & \textbf{Strong perspectivism} (model full annotation distribution; condition predictions on annotator characteristics) \\
\bottomrule
\end{tabular*}

\vspace{2pt}
\raggedright\scriptsize
\textsuperscript{$\dagger$}The boundary between resolvable and irreducible disagreement is a continuum, not a clean partition. In practice, many annotation instances contain both types simultaneously (\cite{schaekermann_resolvable_2018, plank_problem_2022}). The categories above are analytic distinctions that help guide methodological choices, not ontological claims about the nature of any specific disagreement.

\vspace{0.8cm}
\captionof{table}{Weak vs.\ Strong Perspectivism\textsuperscript{$\ddagger$}}
\label{tab:weak-strong}
\vspace{0.2cm}
\begin{tabular*}{\linewidth}{@{\extracolsep{\fill}} p{0.2cm} p{4cm} p{10cm} p{8cm}}
\toprule
& & \textbf{Weak Perspectivism} & \textbf{Strong Perspectivism} \\
\midrule
1. & Epistemological assumption 
  & A correct answer exists; some annotators approximate it better 
  & Multiple valid answers coexist; correctness is group-relative \\[4pt]
2. & Prediction target 
  & Single aggregated label 
  & Distribution over labels or per-group predictions \\[4pt]
3. & Disagreement is treated as 
  & Noise to reduce 
  & Signal to model \\[4pt]
4. & Typical methods 
  & Reliability weighting, Dawid--Skene, item-response models (\cite{raykar_learning_2010, uma_learning_2021})
  & Multi-task heads, soft-label training (\cite{mostafazadeh_davani_dealing_2022, fornaciari_beyond_2021}) \\[4pt]
5. & Appropriate for 
  & Factual coding tasks (entity recognition, event detection, bill status) 
  & Evaluative tasks (sentiment, tone, framing, toxicity) \\[4pt]
6. & Risk if misapplied where the \textit{other} approach is warranted 
  & Compresses legitimate disagreement into a single number; privileges whichever group dominates the annotator pool 
  & Adds unnecessary complexity to factual tasks; may model noise as if it were perspective \\[4pt]
7. & Empirical evidence 
  & Standard in supervised NLP; well-validated for factual annotation 
  & Models trained on disaggregated labels match or outperform majority-vote baselines when evaluated against human distributions (\cite{uma_learning_2021, fornaciari_beyond_2021, pavlick_inherent_2019}) \\
\bottomrule
\end{tabular*}

\vspace{2pt}
\raggedright\scriptsize
\textsuperscript{$\ddagger$}The distinction between weak and strong perspectivism was introduced by \citet{cabitza_perspectivist_2023}. Strong perspectivism does not deny that ground truth exists for factual tasks; it holds that for evaluative tasks the annotation distribution \textit{is} the phenomenon of interest, and collapsing it discards exactly the variation that matters.

\end{table}
\end{landscape}


This paper therefore introduces two contributions: (1) it brings perspectivism from 
text into the visual domain in political science by building a classifier that conditions 
predicted  sentiment on audience identity rather than treating it 
as a fixed property of the image, and (2) it moves beyond single-axis 
conditioning by modeling multiple dimensions of social location 
simultaneously and showing that which dimension drives evaluative 
divergence varies across images. Both follow from the premise that political 
judgment is not independent of the judge. And therefore, identity-triggering evaluations are unavoidably filtered through the group identities they carry, 
which shape the interpretive frames they bring to any given political image.\footnote{The argument likely extends to any visual content 
that engages social or political identity, not only images explicitly 
labeled as political. An image of a neighborhood, a family, or a 
workplace can activate group-based evaluative frames just as readily as 
an image of a protest. What matters is whether the content touches on 
lines of social division. Images that depict politically inert subject 
matter (landscapes, nature morte, abstract patterns) are unlikely to 
produce structured disagreement of this kind.}

\section*{Perspectivist Visual Political Sentiment (PVPS)}

This shift from content-centered prediction to audience-conditioned evaluation requires extending computer vision beyond detecting visual content only. Computer vision has made remarkable progress in political science for tasks 
where the goal is to identify what is in an image and how it is politically 
relevant, including protest framing (\cite{won2017protest,torres_framing_2021}), face 
recognition (\cite{joo_automated_2015}), demographic classification 
(\cite{karkkainen_fairface_2021}), and measurement of crowd size and 
violence (\cite{sobolev_news_2020, steinert_threlkeld_violence_2022}).

\begin{figure}[htbp]
\vspace{0.5cm}    
    \centering
    \includegraphics[width=0.8\textwidth]{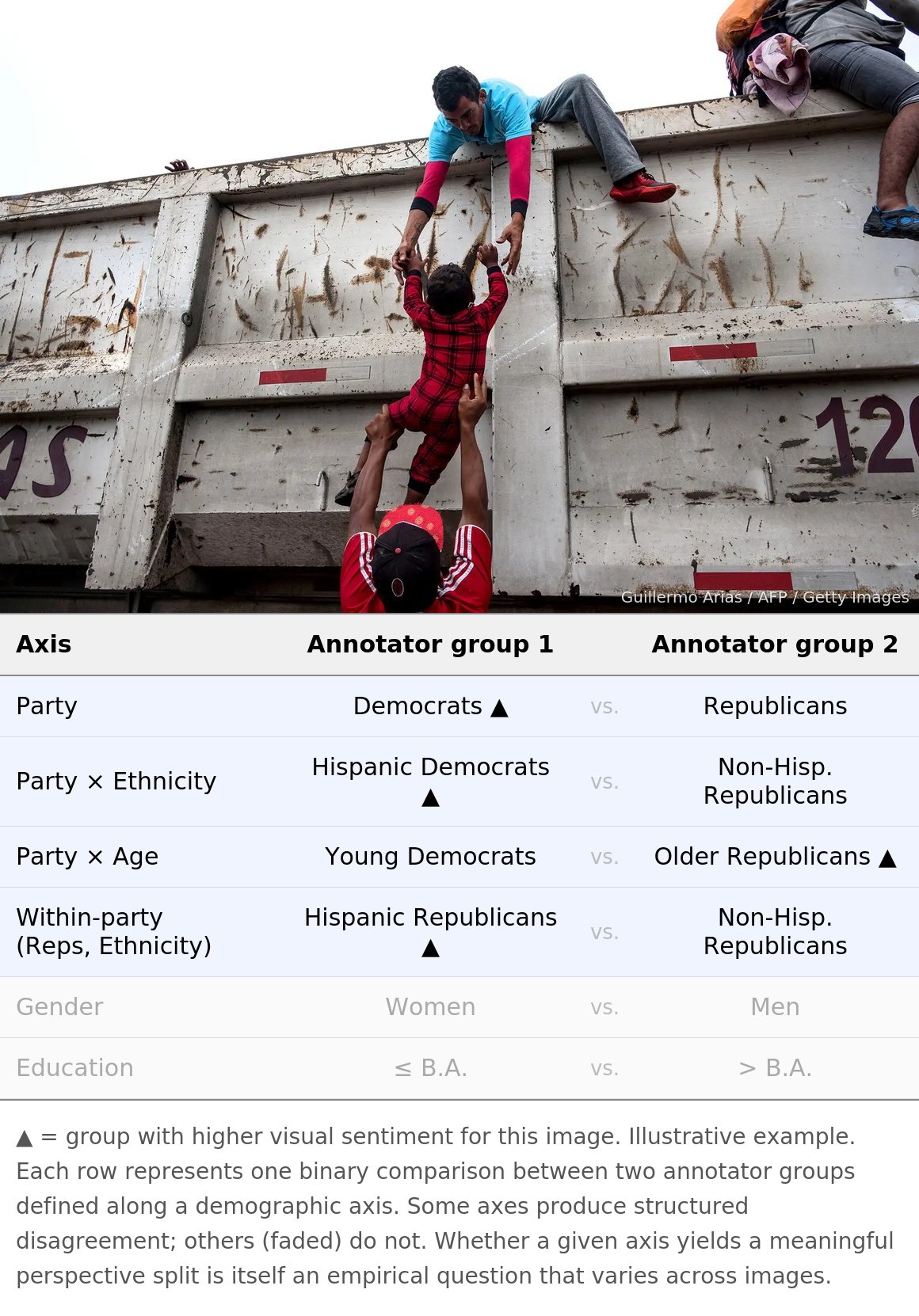}
    \caption{Illustration of how PVPS classification works. The same image 
    yields structured disagreement across certain demographic axes but not 
    others.}
    \label{fig:example}
\vspace{0.5cm}    
\end{figure}

This ground is solid because object recognition tasks are fundamentally 
about factual coding (e.g., whether a crowd is present in an image), and 
these tasks tend to produce substantially higher inter-annotator agreement 
than evaluative ones.\footnote{Even factual image labeling is not free of 
systematic disagreement. \citet{northcutt_pervasive_2021} found approximately 
6\% systematic label errors in the ImageNet validation set, and 
\citet{beyer_imagenet_2020} introduced corrected labels after finding that 
many class boundaries are inherently ambiguous.} The frontier now lies in 
evaluative tasks where the question is not \textit{what} the image contains 
but \textit{what it means for different people}. 

Figure~\ref{fig:example} illustrates that evaluative disagreement over the same image can emerge along some demographic dimensions but not others. Depending on which identity axis we condition on, a photograph may appear politically divisive, culturally polarized, or largely consensual, meaning that majority-vote labels discard meaningful structure in audience response. Experimental and observational research similarly shows that visual framing can shift political attitudes (\cite{soroka_impact_2016}), that media outlets systematically encode events through different visual strategies (\cite{torres_framing_2021, gasparyan_media_2025}), and that identity-relevant visual cues activate group-specific interpretations (\cite{hiaeshutter_rice_cued_2023}). Modeling these dimensions directly therefore makes it possible to see which lines of disagreement are actually activated by a given image and to measure visual persuasion effects at a more granular level than conventional image-level labels allow.

\citet{gasparyan_decoding_2024} apply this logic to political images 
directly, training a multi-task model with party-specific classification 
heads on immigration imagery. This work establishes that perspectivist 
visual sentiment analysis is feasible and informative. But it, like the 
cross-domain work reviewed above, fixes a single political dimension in 
the model architecture.

\citet{williams2026conservatives} continue this perspectivist logic and show that annotation of political images is sensitive to partisan identity and to demographic characteristics such as gender. Regression results on protest mobilization shift depending on whether labels come from Democrats or Republicans, men or women.

The present work borrows from both of these contributions and develops PVPS classifier that takes the next step:

\begin{itemize}
\item The model does not commit to a single social dimension. It computes evaluative divergence along multiple axes simultaneously (partisan, ideological, and demographic) and across their intersections, e.g., young Democrats against old Republicans, high-education Democrats against low-education Democrats, and so on.\footnote{Where the boundary falls on continuous dimensions like age or education is not imposed a priori but searched over empirically, because the location of the fault line is itself part of what needs to be estimated.}
\item  The classifier then learns to predict, from the visual content of the image alone, which side of each divide a given audience segment will fall on. 
\item The result is a tool that can assess whether an image  evokes divergent reactions, along which axes, how confidently the direction of divergence can be predicted from visual content, and whether the fault lines run along partisan, demographic, or intersectional divisions.
\end{itemize}

\section*{Empirical Strategy}
\subsection*{PVPS Model Architecture}

The most intuitive automated approach today would be to 
pass each image through a large language model and ask it to simulate 
audience reactions: ``How would a young Democrat feel about this image? How 
would an old Republican?'' But LLMs tend to encode a \textit{default perspective} 
that reflects the demographic composition of their training 
data (\cite{santurkar_whose_2023}) and embedded stereotypes. When asked to simulate the reactions of 
specific demographic groups, they can approximate some patterns of opinion 
variation (\cite{argyle_out_2023}), but the fidelity is uneven, and for 
visual political content there are few empirical benchmarks tying LLM 
output to how real people actually responded (\cite{egami_using_2024}).

The classifier developed here learns from what different groups \textit{actually} 
reported thinking, which is real evaluative divergence measured in survey data and it generalizes that pattern to unseen images. 
I now describe how this classifier is built.

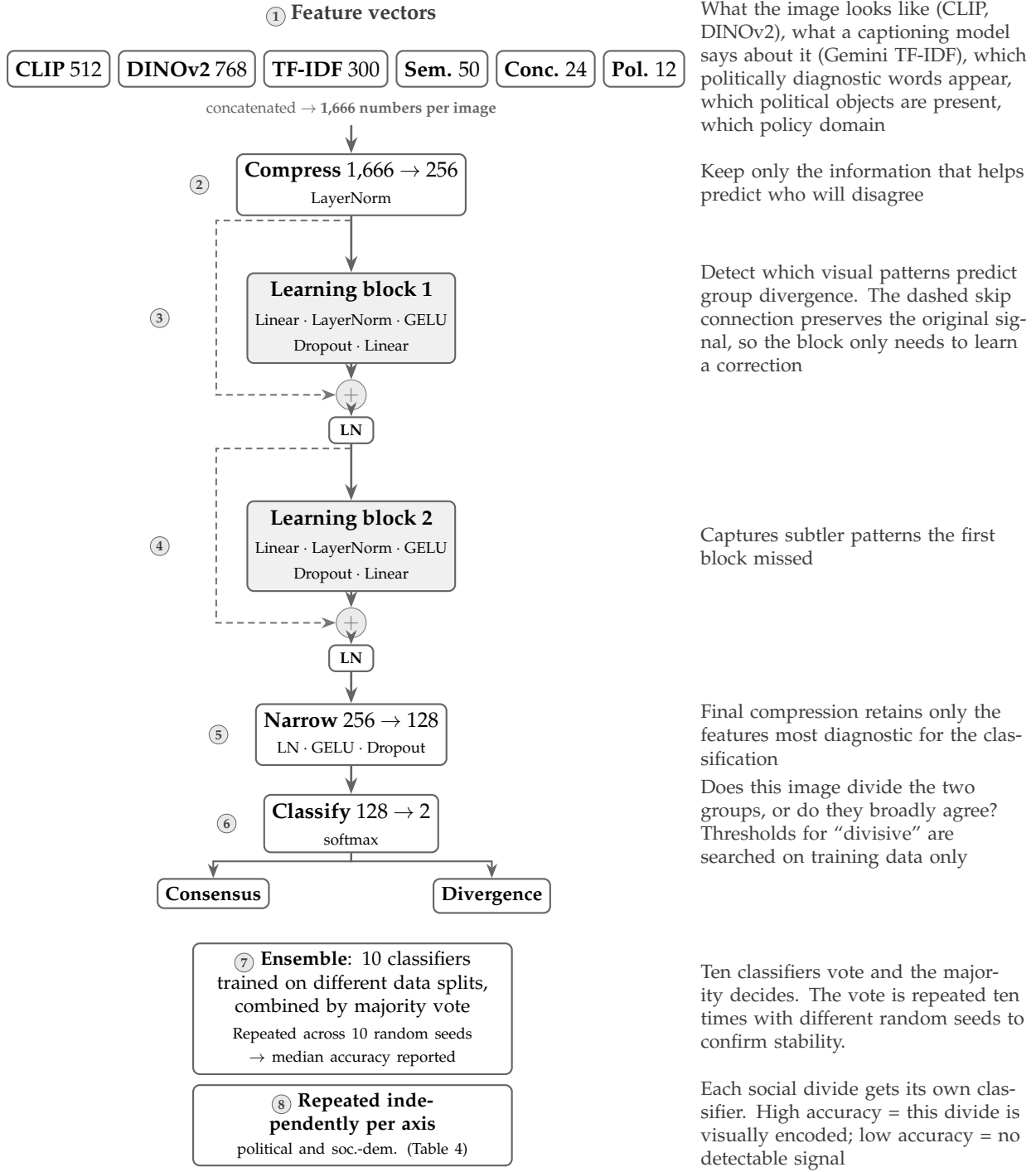
\begin{figure}[H]
\vspace{0.5cm}
\centering
\begin{tikzpicture}[
    >=Stealth,
    scale=0.88, every node/.style={scale=0.88},
    block/.style={
      rectangle, rounded corners=3pt, line width=0.8pt,
      font=\small, align=center, inner sep=4pt,
      draw=medgr, fill=white,
    },
    grayblock/.style={
      block, fill=litegr!20,
    },
    addnode/.style={
      circle, draw=skipgr, fill=skipgr!15,
      inner sep=1.5pt, font=\small\bfseries, text=skipgr,
    },
    ann/.style={font=\footnotesize, text=darkgr, align=left, text width=6cm, anchor=west},
    slab/.style={font=\small\bfseries, text=darkgr},
]

\coordinate (guide) at (11.5, 0);

\node[block, minimum width=1.2cm, minimum height=0.65cm] (clip) at (0,0)
      {\textbf{CLIP} 512};
\node[block, minimum width=1.4cm, minimum height=0.65cm, right=0.12cm of clip]
      (dino) {\textbf{DINOv2} 768};
\node[block, minimum width=1.3cm, minimum height=0.65cm, right=0.12cm of dino]
      (tfidf) {\textbf{TF-IDF} 300};
\node[block, minimum width=0.9cm, minimum height=0.65cm, right=0.12cm of tfidf]
      (sem) {\textbf{Sem.} 50};
\node[block, minimum width=0.9cm, minimum height=0.65cm, right=0.12cm of sem]
      (con) {\textbf{Conc.} 24};
\node[block, minimum width=0.9cm, minimum height=0.65cm, right=0.12cm of con]
      (pol) {\textbf{Pol.} 12};

\node[slab, above=0.25cm of $(clip.north)!0.5!(pol.north)$]
      (feat_lab) {\cmark{1} Feature vectors};
\node[font=\scriptsize, text=medgr, below=0.1cm of $(clip.south)!0.5!(pol.south)$]
      (feat_ann) {concatenated $\to$ \textbf{1,666 numbers per image}};

\node[ann] at (guide |- clip.center)
      {What the image looks like (CLIP, DINOv2), what a captioning model says about it (Gemini TF-IDF), which politically diagnostic words appear, which political objects are present, which policy domain};

\node[block, minimum width=2.5cm, minimum height=1cm, below=0.45cm of feat_ann]
      (proj) {\textbf{Compress} 1{,}666 $\to$ 256\\{\scriptsize LayerNorm}};
\draw[->, line width=1pt, color=medgr] (feat_ann.south) -- (proj.north);
\node[slab, left=0.3cm of proj] {\cmark{2}};

\node[ann] at (guide |- proj.center)
      {Keep only the information that helps predict who will disagree};

\node[grayblock, minimum width=3.4cm, minimum height=1.4cm, below=0.9cm of proj]
      (res1) {\textbf{Learning block 1}\\[1pt]
      {\scriptsize Linear $\cdot$ LayerNorm $\cdot$ GELU}\\
      {\scriptsize Dropout $\cdot$ Linear}};
\draw[->, line width=1pt, color=medgr] (proj.south) -- (res1.north);

\node[addnode, below=0.25cm of res1] (add1) {$+$};
\draw[->, line width=1pt, color=medgr] (res1.south) -- (add1.north);
\draw[->, line width=0.8pt, color=skipgr, densely dashed]
    ($(proj.south)+(0,-0.08)$) -| ($(res1.west)+(-0.55,0)$) |- (add1.west);

\node[block, minimum width=0.8cm, minimum height=0.3cm,
      below=0.1cm of add1, font=\scriptsize\bfseries] (ln1) {LN};
\draw[->, line width=0.8pt, color=medgr] (add1.south) -- (ln1.north);

\node[slab, left=1.1cm of res1] {\cmark{3}};

\node[ann] at (guide |- res1.center)
      {Detect which visual patterns predict group divergence. The dashed skip connection preserves the original signal, so the block only needs to learn a correction};

\node[grayblock, minimum width=3.4cm, minimum height=1.4cm, below=0.9cm of ln1]
      (res2) {\textbf{Learning block 2}\\[1pt]
      {\scriptsize Linear $\cdot$ LayerNorm $\cdot$ GELU}\\
      {\scriptsize Dropout $\cdot$ Linear}};
\draw[->, line width=1pt, color=medgr] (ln1.south) -- (res2.north);

\node[addnode, below=0.25cm of res2] (add2) {$+$};
\draw[->, line width=1pt, color=medgr] (res2.south) -- (add2.north);
\draw[->, line width=0.8pt, color=skipgr, densely dashed]
    ($(ln1.south)+(0,-0.08)$) -| ($(res2.west)+(-0.55,0)$) |- (add2.west);

\node[block, minimum width=0.8cm, minimum height=0.3cm,
      below=0.1cm of add2, font=\scriptsize\bfseries] (ln2) {LN};
\draw[->, line width=0.8pt, color=medgr] (add2.south) -- (ln2.north);

\node[slab, left=1.1cm of res2] {\cmark{4}};

\node[ann] at (guide |- res2.center)
      {Captures subtler patterns the first block missed};

\node[block, minimum width=2.5cm, minimum height=1cm, below=0.5cm of ln2]
      (head) {\textbf{Narrow} 256 $\to$ 128\\{\scriptsize LN $\cdot$ GELU $\cdot$ Dropout}};
\draw[->, line width=1pt, color=medgr] (ln2.south) -- (head.north);
\node[slab, left=0.3cm of head] {\cmark{5}};

\node[ann] at (guide |- head.center)
      {Final compression retains only the features most diagnostic for the classification};

\node[block, minimum width=2cm, minimum height=0.8cm, below=0.45cm of head]
      (out) {\textbf{Classify} 128 $\to$ 2\\{\scriptsize softmax}};
\draw[->, line width=1pt, color=medgr] (head.south) -- (out.north);

\node[block, minimum width=1.6cm, minimum height=0.5cm,
      below left=0.4cm and -0.1cm of out, font=\footnotesize\bfseries] (c0) {Consensus};
\node[block, minimum width=1.6cm, minimum height=0.5cm,
      below right=0.4cm and -0.1cm of out, font=\footnotesize\bfseries] (c1) {Divergence};
\draw[->, line width=0.8pt, color=medgr] (out.south) -- ++(0,-0.15) -| (c0.north);
\draw[->, line width=0.8pt, color=medgr] (out.south) -- ++(0,-0.15) -| (c1.north);

\node[slab, left=0.3cm of out] {\cmark{6}};

\node[ann] at (guide |- out.center)
      {Does this image divide the two groups, or do they broadly agree? Thresholds for ``divisive'' are searched on training data only};

\node[block, text width=5.5cm, inner sep=4pt,
      below=0.5cm of $(c0.south)!0.5!(c1.south)$,
      font=\footnotesize, align=center]
      (ens) {\cmark{7} \textbf{Ensemble}: 10 classifiers trained on different
      data splits, combined by majority vote\\[2pt]
      {\scriptsize Repeated across 10 random seeds $\to$ median accuracy reported}};

\node[ann] at (guide |- ens.center)
      {Ten classifiers vote and the majority decides. The vote is repeated ten times with different random seeds to confirm stability.};

\node[block, text width=5.5cm, inner sep=4pt,
      below=0.15cm of ens, font=\footnotesize, align=center]
      (ax) {\cmark{8} \textbf{Repeated independently per axis}\\
      {\scriptsize political and soc.-dem.
      (Table~\ref{tab:axes})}};

\node[ann] at (guide |- ax.center)
      {Each social divide gets its own classifier. High accuracy = this divide is visually encoded; low accuracy = no detectable signal};

\end{tikzpicture}
\caption{Pipeline of PVPS classifier architecture.}
\label{fig:pipeline}
\vspace{0.5cm}
\end{figure}

\begin{description}

\item[\cmark{1} Feature vectors.]
Each image is translated into a numerical vector
that the classifier can process.\footnote{For a political-science-oriented introduction to how pretrained models process images, see \cite{torres2022learning}.}
Each image is passed through a set of pretrained models that extract different
aspects of its content and encode them as numbers.

I use six extractors because they capture complementary information.\footnote{An ablation study (Appendix~\ref{sec:app_ablation}) evaluates the contribution of each encoder to classifier performance.}
\textbf{CLIP} (512 numbers) and \textbf{DINOv2} (768 numbers) both produce
numerical summaries of what an image looks like, but they learn in different
ways. CLIP was trained on hundreds of millions of image--caption pairs, so it
encodes visual content together with the symbolic associations carried by
language (e.g., a raised fist mapped close to ``protest''). DINOv2 was trained
on images alone, without any text, so it captures texture, shape, and spatial
layout independently of verbal framing.\footnote{See Appendix~\ref{sec:app_encoders} for encoder details.}
\textbf{Gemini captions} (300 numbers) add an explicit textual layer\footnote{A large language model (Gemini) generates a structured natural-language description of each image
(e.g., ``a group of people holding signs in front of a government building''). The 300 dimensions correspond to the 300 most informative words in the caption
vocabulary.}
and these descriptions are converted into numerical vectors via
TF-IDF.\footnote{TF-IDF (Term Frequency--Inverse Document Frequency) assigns
high weight to words that are frequent in a given caption but rare across the
corpus. The vocabulary is fitted on the training split only to prevent leakage.} \textbf{Semantic features} (50 numbers) are binary indicators of whether
specific predictive words appear in the caption, selected per axis and per seed
by a chi-squared test on training data.\footnote{For example, if the word
``protest'' is strongly associated with one class, the corresponding feature is
1 when that word appears in the caption and 0 otherwise. Unlike TF-IDF, which
weights word frequency, these features record presence or absence of specific
words selected for their predictive power.}
\textbf{Concept features} (24 numbers) record the presence or absence of
predefined political concepts (protest signs, police, politicians, national
symbols, etc.) drawn from the political image
literature.\footnote{Refined during hyperparameter search to retain only
concepts that contributed to classification accuracy. See Appendix~\ref{sec:app_concepts}.}
\textbf{Political topic features} (12 numbers) indicate which policy domain
the image belongs to (immigration, guns, LGBTQ+, climate, etc.), detected
from the Gemini description by keyword matching.
These six vectors are concatenated into a single 1,666-number fingerprint per image.

\item[\cmark{2} Projection.]
These 1,666 numbers contain redundancies and noise. The projection layer
compresses them to 256 through a linear transformation followed by layer
normalization, forcing the network to retain only the information that matters
for predicting group disagreement and discard what does not.

\item[\cmark{3} Residual Block~1.]
The network begins learning which combinations of visual features predict group disagreement. The block transforms the signal through two linear layers with normalization, a nonlinear activation function,\footnote{A function that allows the network to capture relationships that are not simple straight lines. Without it, stacking multiple layers would be equivalent to a single linear transformation.} and dropout.\footnote{During training, a random fraction of connections is temporarily switched off at each step. This forces the network to spread useful information across many features and prevents it from relying too heavily on any one of them.} A skip connection (the dashed line in Figure~\ref{fig:pipeline}) adds the block's input directly to its output, so the block only needs to learn a correction to the signal it receives.\footnote{Skip connections make deeper networks easier to train because adding a block can never make the representation worse. In the worst case, the block learns to output zeros and passes the input through unchanged.}

\item[\cmark{4} Residual Block~2.]
A second block with identical architecture refines the representation
further, picking up patterns that the first block
missed.\footnote{Stacking two blocks improved accuracy compared to one;
adding a third provided no further gain for data of this size.}

\item[\cmark{5} Output head.]
The 256-dimensional representation is compressed to 128 through another linear
layer with normalization, activation, and dropout, retaining only the features
most diagnostic of the final classification.

\item[\cmark{6} Classification.]
A final linear layer maps the 128 numbers to class probabilities via
softmax. For the standard binary task the output is two probabilities:
\textit{consensus} (both groups rate the image similarly) versus
\textit{divergence} (one group rates it substantially higher). The image is
assigned to whichever class receives the higher probability.\footnote{For
multiclass axes (more than two groups), the output layer produces one
probability per class and the ensemble averages the full probability
distributions across members before assigning the most probable class.}

For each axis, the pipeline defines what counts as ``divergent'' by computing
the evaluative gap between groups per image across respondents in the training
split and searching over candidate thresholds to find the split that best
separates consensual from divisive images.\footnote{For continuous variables like age
and education, the group boundary itself is part of the search: ``young
versus old'' could mean splitting at 35, 40, or 45, and the pipeline retains
whichever definition produces the best class separation on training
data. All threshold and boundary decisions are made on the training
split only; test images are never used.}

\item[\cmark{7} Ensemble.]
No single random split of the data should determine the result. Ten copies of the classifier are trained on different partitions of the data\footnote{Each partition is stratified, meaning both the training and validation sets preserve the proportion of positive and negative examples found in the full dataset.} and their predictions are combined by majority vote (for binary axes) or by averaging their probability outputs (for multiclass axes). This procedure repeats across ten independent random seeds.\footnote{A random seed is a number that fixes the sequence of all pseudo-random decisions in the pipeline, such as how data are split and how network weights are initialized. Changing the seed produces a different run of the same experiment. Repeating across multiple seeds tests whether results are stable or depend on a particular sequence of random choices.} The reported accuracy is the median of the minimum per-class accuracy across seeds, a metric that guards against both lucky splits and classifiers that achieve high overall accuracy by correctly predicting only the more frequent class.

\item[\cmark{8} Repeated independently per axis.]
The full pipeline runs separately for each axis and intersection listed in Table~\ref{tab:axes}. Each axis gets its own group boundaries and trained classifier. High accuracy on a given axis means the visual content of the images carries enough information to predict how evaluations will diverge along that social divide. Low accuracy means no such signal is detectable.

\end{description}

\subsection*{Data Sources}

The classifier draws on five pools of annotated political images that differ in annotation density, demographic coverage, and the role they play in the pipeline (Table~\ref{tab:data_comparison}).

\begin{table}[htbp]
\vspace{0.5cm}
\centering
\small
\caption{Data sources.}
\label{tab:data_comparison}
\begin{tabular}{lcccp{7cm}}
\toprule
\textbf{Source} & \textbf{Imgs} & \textbf{Resp.} & \textbf{Annot.} & \textbf{Topics} \\
\midrule
Wave 1\textsuperscript{a}     & 356     & 2{,}089   & 16.7K  & Immigration \\
Waves 2+3\textsuperscript{b}  & 852     & 4{,}327   & 54.9K  & Immigration \\
Wave 4\textsuperscript{c}     & 412     & 1{,}248   & 27.1K  & Immigration, guns, LGBT+, abortion, January~6 \\
Webb Williams et al.\ (2026)\textsuperscript{*}         & 7{,}543 & ${\sim}$1--3/img & 19.8K  & Immigration, guns, climate, gender, politics, racial justice, human rights \\
\bottomrule
\end{tabular}

\vspace{3pt}
\raggedright\scriptsize
\textsuperscript{a}Gasparyan and Sirotkina (2025); Lucid, April 2022. A filtered subset of Wave~2 respondents and images, published in \textit{PLOS ONE}. \\
\textsuperscript{b}Gasparyan and Sirotkina (2024); two Lucid surveys (April 2022 and March 2024). \\
\textsuperscript{c}Prolific, 2024--2025. \\
\textsuperscript{*}Webb Williams et al.\ (2026); Mechanical Turk. Supplementary data, incorporated via semi-supervised self-training (see clarification below). \\
\vspace{0.5cm}
\end{table}

The primary dataset combines four survey waves (2022--2025, Prolific and Lucid), totaling 1,264 unique images and approximately 82,000 image-by-respondent observations from 5,575 U.S.\ adults. Waves~1--3 covered immigration \citep{gasparyan_media_2025, gasparyan_decoding_2024}; Wave~4 added guns, LGBT+ rights, abortion, and January~6. Each respondent rated images on a 7-point attitude scale and completed a demographic battery (party identification, ideology, feeling thermometers, age, gender, education, income, Hispanic ethnicity). Each image was evaluated by a median of roughly 65 respondents; per-image group comparisons require at least five respondents per group. A supplementary dataset from \citet{williams2026conservatives} provides 7,543 images rated by MTurk workers on discrete emotions across six policy domains, mapped onto the same 7-point scale via quantile matching (r=0.917; partisan-gap sign agreement 92.2\%).\footnote{Because these data contain only two to three annotators per image, the pipeline incorporates them only through semi-supervised self-training \citep{yarowsky_unsupervised_1995, lee_pseudolabel_2013}, pseudo-labeling confident cases (predicted probability $\geq$ 0.75), with a diagnostic that disables supplementary data if source-classification accuracy exceeds 85\%.} I surveyed several additional datasets, but they lacked the annotator structure necessary for perspectivist classification.\footnote{I surveyed four candidate image datasets that provide individual-level evaluative ratings of political images: \citet{todorov_inferences_2005, todorov_face_2017}, who collect trait judgments from multiple raters but record no political attributes; \citet{casas_images_2019}, who collect emotion ratings but with only two to three annotators per image and no demographic battery; and \citet{webb_williams_images_2020}, who record annotator partisanship and gender but again with minimal annotation density per image. Because the perspectivist paradigm \citep{basile_we_2021, cabitza_perspectivist_2023} is not yet standard practice, most political image datasets either aggregate annotations into a single label or discard individual-annotator records entirely, making them unusable for group-conditional training. Only the five sources described above met the pipeline's requirements, which is individual-level ratings linked to respondent-level political and demographic attributes, with sufficient annotation density per image to estimate group-level evaluative differences.}

\begin{table}[H]
\vspace{0.5cm}
\centering
\small
\caption{Classification axes. Each row defines a binary comparison 
estimated independently by the full pipeline.}
\label{tab:axes}

\begin{tabular}{p{3.5cm} p{9.5cm}}
\toprule
\textbf{Block} & \textbf{Examples} \\
\midrule
\textit{Political} & 
  Democrat vs.\ Republican \newline
  Liberal vs.\ Conservative \newline
  Pro-Democrat vs.\ Pro-Republican (feeling thermometer) \\[6pt]
\textit{Demographic} & 
  Female vs.\ Male \newline
  Hispanic vs.\ non-Hispanic \newline
  Young vs.\ Old; High-ed.\ vs.\ Low-ed.; High-inc.\ vs.\ Low-inc.\textsuperscript{a} \\[6pt]
\textit{Cross-political $\times$ demographic} & 
  Young Democrat vs.\ Old Republican (and reverse) \newline
  Female Liberal vs.\ Male Conservative (and reverse) \newline
  \dots both directions for every political--demographic pair\textsuperscript{b} \\[6pt]
\textit{Within-political $\times$ demographic} & 
  Young Democrat vs.\ Old Democrat \newline
  Female Republican vs.\ Male Republican \newline
  \dots each demographic split within each political group\textsuperscript{c} \\[6pt]
\textit{Same-demographic, cross-political} & 
  Young Democrat vs.\ Young Republican \newline
  Female Liberal vs.\ Female Conservative \newline
  \dots same demographic, different political identity\textsuperscript{d} \\
\bottomrule
\end{tabular}

\vspace{3pt}
\raggedright\scriptsize
\textsuperscript{a}For continuous demographics (age, education, income), 
the group boundary is searched empirically on training data; the best 
split (into 2, 3, or 4 groups) is selected automatically. \\
\textsuperscript{b}Cross-political axes test whether evaluative 
divergence is detectable when the comparison crosses a political 
divide and a demographic boundary simultaneously. Both directions are 
tested, yielding 36 axes across the three political anchors (party, 
ideology, thermometer). \\
\textsuperscript{c}Within-political axes are diagnostic nulls: if the 
classifier cannot detect demographic divergence \textit{within} a 
political group using the same features, the visual signal is 
political, not demographic. 38 axes total. \\
\textsuperscript{d}Same-demographic axes hold the demographic constant 
and vary only political identity, isolating the partisan component. 
30 axes total. \\
Total: 112 axes.
\vspace{0.5cm}
\end{table}

Table~\ref{tab:axes} lists the classification axes. The first block captures the partisan divide at varying intensities; the second crosses party with a demographic attribute; the third isolates demographics within each party. This structure reflects stable, overlapping cleavages in American politics, where partisan identity has grown increasingly aligned with ideology and social identity \citep{mason_uncivil_2018, iyengar_origins_2019, boxell_cross-country_2022}, while age, education, gender \citep{sides_identity_2018, schaffner_understanding_2018}, and ethnicity \citep{hajnal_why_2011, reny_vote_2019} produce meaningful within-party variation. The within-party axes serve as a diagnostic. If the classifier detects evaluative divergence along a demographic dimension within a party using the same features that work across parties, the signal is at least partly demographic rather than purely partisan.

\section*{Results}

\subsection*{Overall Classification Performance}

\begin{figure}[H]
    \centering
    \hspace*{-1.0cm}%
    \includegraphics[width=1.1\textwidth]{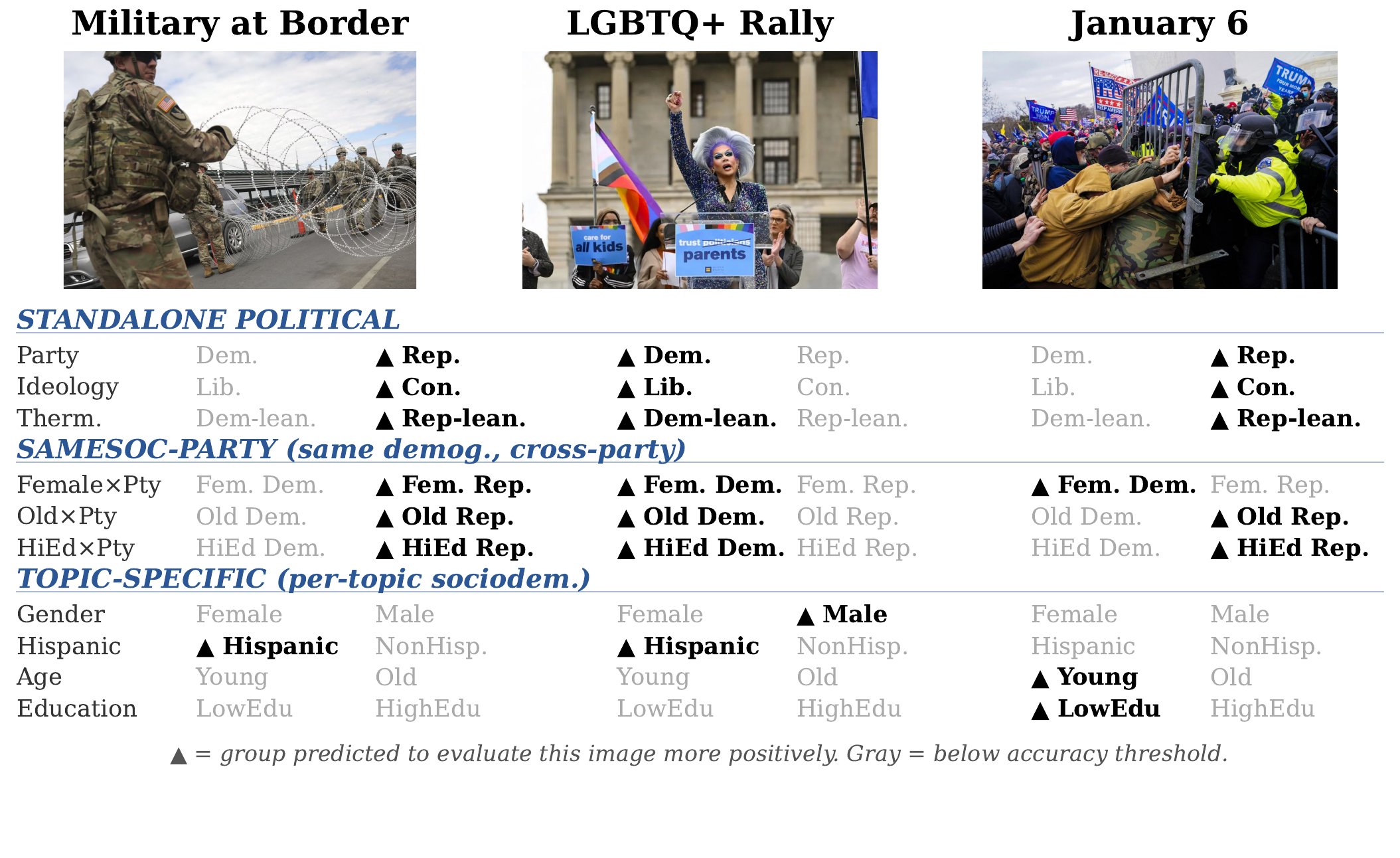}
        \caption{Evaluative profiles for three held-out images.}
            \vspace{-1.5cm}    
    \label{fig:eval_prof}
    \vspace{0.5cm}    
\end{figure}

Figure~\ref{fig:eval_prof} displays evaluative profiles for three 
held-out test images. Full profiles across all 112 axes are reported 
in Appendix~\ref{sec:app_profiles}.
On the military-at-border and January~6 images, most political axes align toward Republican, Conservative, and Republican-leaning respondents, while the LGBTQ+ rally reverses this pattern across nearly all political dimensions. Same-sociodemographic cross-party comparisons remain active, showing that partisan disagreement persists even when demographic composition is held constant. In most cases, these cross-party axes follow the broader partisan direction, though January~6 produces some localized reversals, particularly among women.

\subsection*{Political Axes, Demographic Axes, and Their Interaction}

\subsubsection*{Political Axes}

The classifier clears the 65\% minimum-class accuracy threshold\footnote{A 65\% minimum-class accuracy threshold is conservative for a task that predicts subjective evaluative divergence compared to objective image content. Comparable visual political classification tasks achieve 59--72\% with far larger datasets and less abstract targets \citep{kosinski_facial_2021, xi_ideology_2020}, and the threshold here applies to minimum per-class accuracy, which is a stricter criterion than overall accuracy and corresponding to a medium effect size ($d \approx 0.5$) \citep{yarkoni_choosing_2017}.} on all three primary political axes (see Appendix~\ref{sec:app_threshold} for discussion of this threshold). Party identification reaches 68.7\% median minimum-class accuracy over ten seeds, ideology reaches 78.9\%, and feeling-thermometer difference reaches 78.5\%.\footnote{Throughout the paper, reported accuracy refers to the median over ten independent random seeds of the minimum per-class accuracy on held-out test images. This is the accuracy on whichever class is harder to predict, which guards against classifiers that achieve high overall accuracy by getting only the more common class right. Accuracy on the other class is higher in every case. Full per-class and per-seed results appear in Appendix~\ref{sec:app_full_results}. All reported numbers use primary data only. The supplementary dataset was disabled automatically by a pre-training source-prediction diagnostic, since the two image pools are visually distinguishable 97\% of the time and a classifier trained on both could learn to separate datasets rather than partisan evaluations (Appendix~\ref{sec:app_source}).}

Same-sociodemographic cross-political axes isolate the partisan component by holding demographics constant and varying only political identity. Female Dem.\ vs.\ Female Rep.\ reaches 82.0\%, High-education Dem.\ vs.\ High-education Rep.\ 77.2\%, Old Liberal vs.\ Old Conservative 79.7\%, and Female Democrat-leaning vs.\ Female Republican-leaning 77.6\%. The partisan signal survives when demographics are equalized. 


\subsubsection*{Sociodemographic Axes}

The image-level classifier fails almost uniformly on sociodemographic axes, with median accuracies between 50\% and 58\% across all thirteen dimensions.  That result is substantively plausible given that political images are constructed to communicate political conflict directly, whereas sociodemographic differences are weaker, more conditional, and often entangled with other respondent characteristics. \footnote{If we randomly split the respondents in each group into two halves and compute the gap separately, the two estimates correlate at only $r = 0.20$ for gender and $r = 0.27$ for age, compared to $r = 0.74$ for party. No classifier can predict a target that is 80\% noise. Second, the direction of sociodemographic gaps reverses across topics (76\% of gun images favoring women, 78\% of LGBTQ+ images favoring men), so a pooled classifier encounters contradictory labels for visually similar content.}

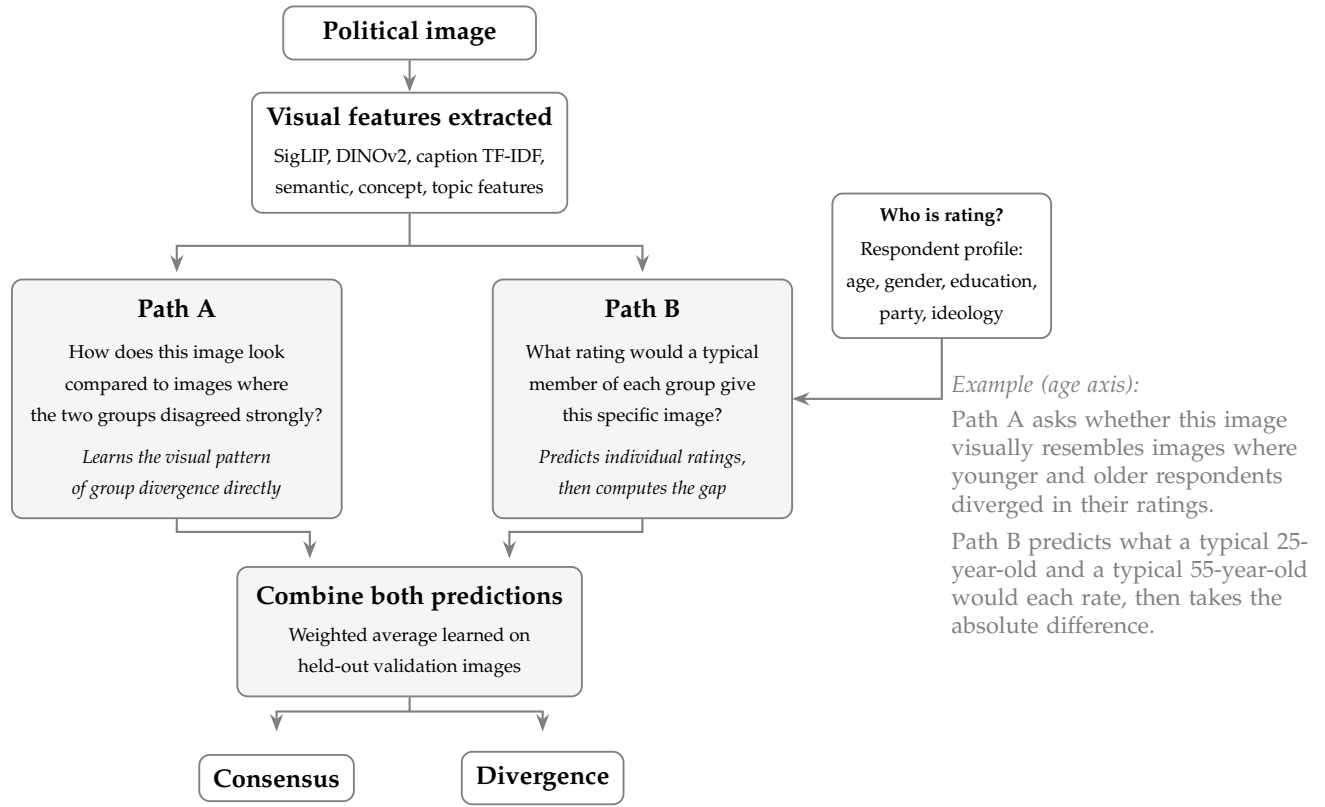
\begin{figure}[H]
\vspace{0.5cm}
\centering
\begin{tikzpicture}[
    >=Stealth,
    scale=0.88, every node/.style={scale=0.88},
    box/.style={
      rectangle, rounded corners=4pt, line width=0.7pt,
      draw=black!50, fill=white,
      font=\small, align=center, inner sep=6pt,
      minimum width=3.8cm
    },
    pathbox/.style={
      rectangle, rounded corners=4pt, line width=0.7pt,
      draw=black!50, fill=black!4,
      font=\small, align=center, inner sep=8pt,
      minimum width=4.5cm
    },
    arr/.style={->, line width=0.8pt, color=black!50}
]

\node[box] (img) at (0, 0)
      {\textbf{Political image}};

\node[box] (feat) at (0, -1.8)
      {\textbf{Visual features extracted}\\[2pt]
       \scriptsize SigLIP, DINOv2, caption TF-IDF,\\
       \scriptsize semantic, concept, topic features};
\draw[arr] (img) -- (feat);

\node[pathbox, minimum height=2.8cm] (pathA) at (-3.5, -5.5)
      {\textbf{Path A}\\[4pt]
       \scriptsize How does this image look\\
       \scriptsize compared to images where\\
       \scriptsize the two groups disagreed strongly?\\[4pt]
       \scriptsize \textit{Learns the visual pattern}\\
       \scriptsize \textit{of group divergence directly}};

\node[pathbox, minimum height=2.8cm] (pathB) at (3.5, -5.5)
      {\textbf{Path B}\\[4pt]
       \scriptsize What rating would a typical\\
       \scriptsize member of each group give\\
       \scriptsize this specific image?\\[4pt]
       \scriptsize \textit{Predicts individual ratings,}\\
       \scriptsize \textit{then computes the gap}};

\draw[arr] (feat.south) -- (0, -3.2) -- (-3.5, -3.2) -- (-3.5, -3.6);
\draw[arr] (0, -3.2) -- (3.5, -3.2) -- (3.5, -3.6);

\node[box, minimum width=2.5cm] (resp) at (8, -3.5)
      {\scriptsize \textbf{Who is rating?}\\[1pt]
       \scriptsize Respondent profile: \\
      \scriptsize age, gender, education,\\
       \scriptsize party, ideology};
\draw[arr] (resp.south) -- (8, -5.5) -- (5.75, -5.5);

\node[pathbox, minimum width=5cm] (combine) at (0, -9)
      {\textbf{Combine both predictions}\\[2pt]
       \scriptsize Weighted average learned on\\
       \scriptsize held-out validation images};

\draw[arr] (pathA.south) -- (-3.5, -7.5) -- (-1.5, -7.5) -- (-1.5, -7.9);
\draw[arr] (pathB.south) -- (3.5, -7.5) -- (1.5, -7.5) -- (1.5, -7.9);

\node[box, minimum width=2cm, font=\small\bfseries] (out0) at (-2, -11.2) {Consensus};
\node[box, minimum width=2cm, font=\small\bfseries] (out1) at (2, -11.2) {Divergence};
\draw[arr] (combine.south) -- (0, -10.2) -- (-2, -10.2) -- (-2, -10.5);
\draw[arr] (0, -10.2) -- (2, -10.2) -- (2, -10.5);

\node[font=\footnotesize, text=black!50, align=left, text width=5.5cm, anchor=north west] at (8, -5)
      {\textit{Example (age axis):}\\[3pt]
       Path A asks whether this image visually resembles images where younger and older respondents diverged in their ratings.\\[4pt]
       Path B predicts what a typical 25-year-old and a typical 55-year-old would each rate, then takes the absolute difference.};

\end{tikzpicture}
\caption{Extended pipeline for sociodemographic axes. The main PVPS classifier (Figure~\ref{fig:pipeline}) fails on sociodemographic dimensions because the per-image evaluative gaps are too noisy. This extension combines two complementary prediction strategies per axis. Path~A predicts the evaluative gap from visual content alone. Path~B predicts individual ratings from image features and respondent attributes, then derives the gap. A weighted combination learned on validation data produces the final classification.}
\label{fig:stack_v2}
\end{figure}

To recover this weaker signal, I extend the main PVPS pipeline (Figure~\ref{fig:pipeline}) with an additional prediction component that conditions on respondent attributes and separates sociodemographic effects from overlapping political structure. Figure~\ref{fig:stack_v2} illustrates the extended pipeline. Path~A is the original image-level classifier and asks whether an image resembles images where one group rated higher than another.\footnote{On political axes this approach works well because the per-image partisan gap is estimated from roughly thirty Democrats and thirty Republicans and is stable enough to learn from. On the age axis, each group (young vs.\ old) may contain only seven to ten respondents per image, the resulting gap is 80\% sampling noise, and Path~A has little to learn. The set of visual encoders fed into Path~A is tailored per axis in a prior ablation study (Appendix~\ref{sec:app_ablation}), because including all six feature sources at once introduces noise that drowns out the weaker sociodemographic signal. For example, the age axis uses SigLIP with DINOv2 and caption features; the education axis uses concept indicators with handcrafted political features. These combinations were fixed before the main evaluation.}
Path~B  learns directly from approximately 93,000 individual ratings, modeling how respondent attributes interact with image content while holding other dimensions constant. It estimates whether a given sociodemographic attribute systematically shifts evaluations of particular types of political imagery while avoiding reliance on noisy group-level averages. At test time, the model compares predicted evaluations for two otherwise similar prototypical respondents who differ only on the target dimension, such as age or education.\footnote{This is analogous to a regression with interaction terms (rating $\sim$ image features $\times$ respondent age), estimated on 93,000 observations rather than on 1,264 noisy per-image means of seven respondents each.} A meta-learner combines both paths, learning how much weight to place on the direct visual signal versus the respondent-conditioned signal for each axis.\footnote{Ridge regression fitted on held-out validation images. On political axes Path~A dominates because the per-image gap is reliable; on sociodemographic axes Path~B contributes more because it extracts signal from individual ratings that the per-image gap is too noisy to preserve. Decision thresholds are calibrated on validation data only, and ten independent random seeds assess stability.}

Under this extended pipeline, eight sociodemographic axes exceed 60\% median minimum-class accuracy. Age reaches 68.9\% ($r = 0.619$), gender 62.0\% ($r = 0.489$), and education 61.0\% ($r = 0.419$), with several within-party and within-ideology age comparisons also producing stable signal.\footnote{The remaining sociodemographic axes fall between 42\% and 59\%. Income produces no signal on any formulation. Hispanic ethnicity axes do not reach meaningful accuracy, consistent with their low split-half reliability ($r = 0.11$). Within-party, within-ideology, and within-thermometer gender and education axes also remain near chance.} These results suggest that social identity does shape evaluations of political imagery, but the visual signal is substantially weaker and more conditional than the partisan signal. Recovering it requires conditioning on overlapping respondent characteristics rather than relying on image-level disagreement alone.

\section*{Model Application}

The classifier was trained and evaluated on survey data mainly collected more or less for 
this study. To test whether it produces meaningful predictions on 
images it has never seen, I apply it to two published datasets and 
ask whether the PVPS evaluative profiles change the substantive 
conclusions those studies report.

\subsection*{Casas and Webb Williams (2019)}

\begin{figure}[H]
\vspace{0.5cm}    
    \centering
    \includegraphics[width=1.1\textwidth]{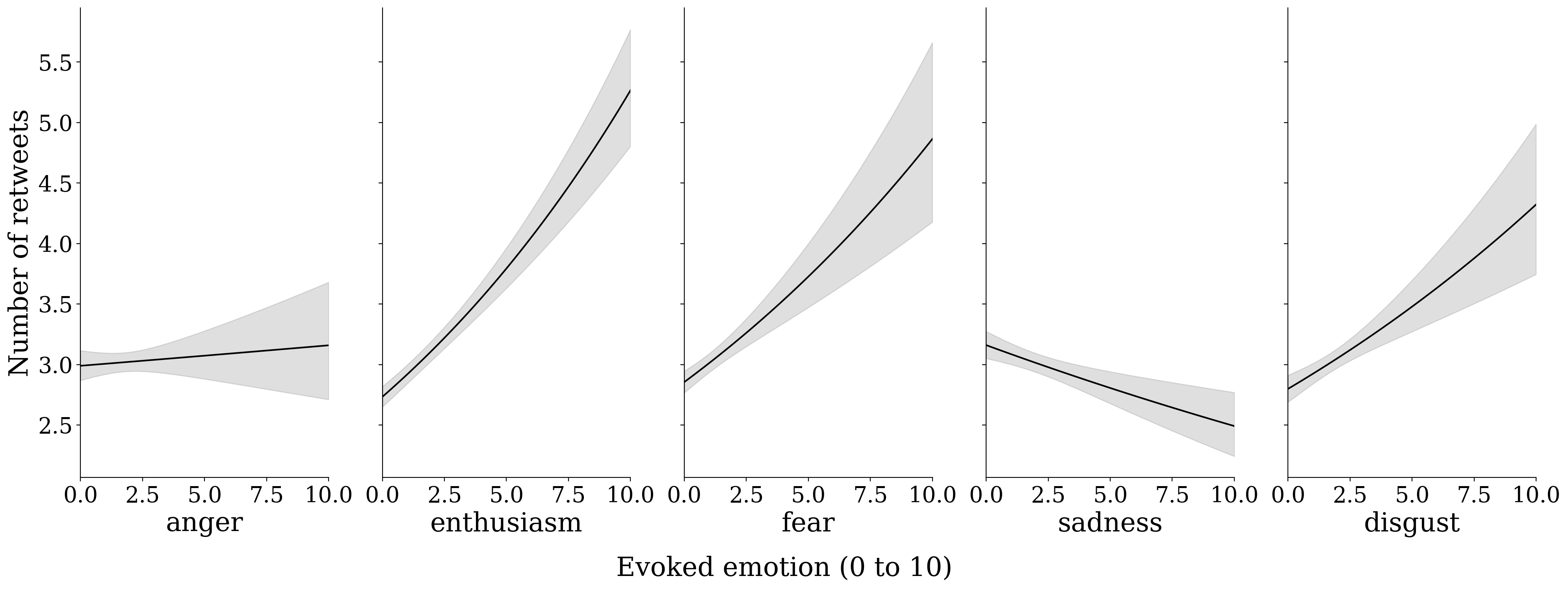}
    \vspace{0.3cm}
    \includegraphics[width=1.1\textwidth]{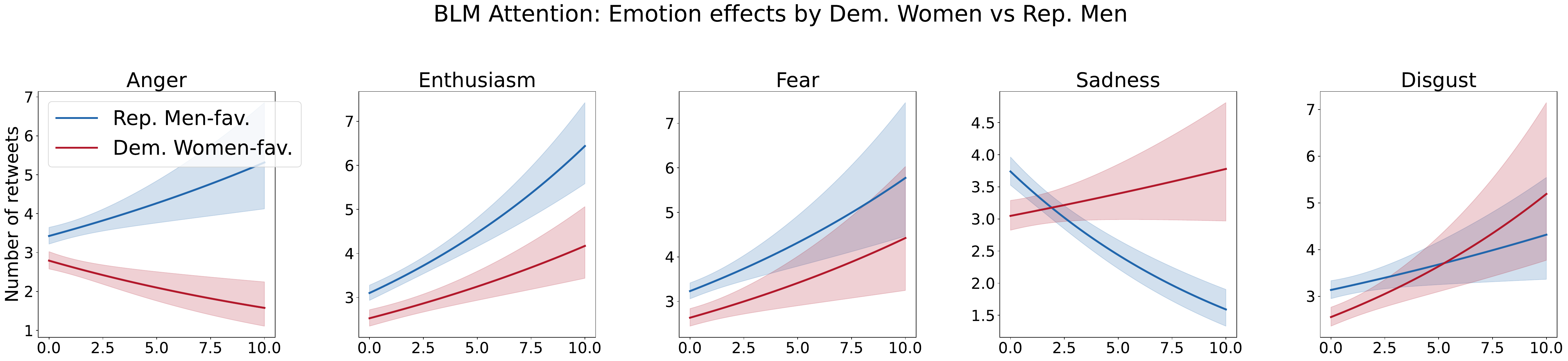}
    \vspace{0.3cm}
    \includegraphics[width=1.1\textwidth]{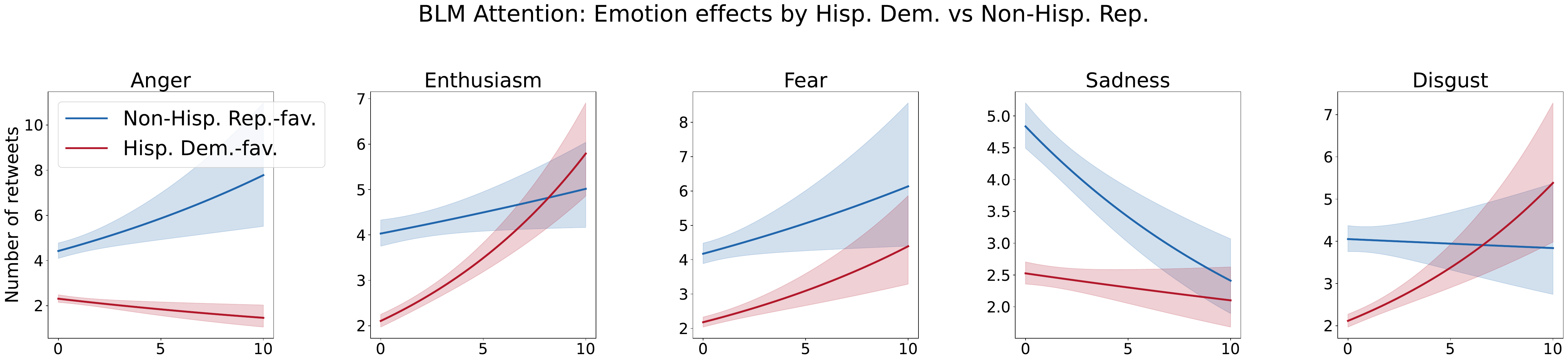}
    \caption{\footnotesize Predicting attention to BLM images over the range of evoked emotions. Top row replicates Casas and Webb Williams (2019, Figure 6), showing predicted retweet counts as a function of emotion intensity (0--10) with 95\% confidence bands. Middle row splits predicted retweets by PVPS score on the Democratic Women vs.\ Republican Men axis (blue = Republican-men-favorable images, red = Democratic-women-favorable). Bottom row splits by the Hispanic Democrat vs.\ Non-Hispanic Republican axis. Diverging lines indicate that the mobilizing effect of a given emotion depends on the partisan-demographic perspective the image conveys. All models are negative binomial with the same controls as the original specification. Confidence bands are computed via the delta method.}
    \label{fig:emotion_heterogeneity}
\vspace{0.5cm}    
\end{figure}

Casas and Webb Williams (2019) collect 150,000 BLM tweets (9,500 images, ShutdownA14, April 2015) and have 1,259 MTurk workers rate each image on five emotions (0--10).\footnote{The top $\sim$950 most-retweeted images were each labeled by five annotators (two undergraduate research assistants and three MTurk workers), with the emotion score averaged across the five. The remaining 8,509 images were labeled by a single annotator drawn from a pool of 1,259 MTurk workers, each of whom could label at most 100 images \citep[367--68]{casas_images_2019}.} Negative binomial regressions predicting retweet count show that enthusiasm and fear increase engagement while sadness decreases it (Figure~\ref{fig:emotion_heterogeneity}, top row, replicated from their data). The emotion scores are averaged across annotators and treated as fixed properties of each image.

I run the same 8,013 images through the PVPS classifier, producing predicted probabilities for each passing axis. I then interact each emotion score with the PVPS probability to test whether the mobilizing effect of a given emotion depends on which group the image favors. The interaction model takes the form

\begin{equation}
\text{retweets} \sim \text{emotion}_j + \text{PVPS}_k + \text{emotion}_j \times \text{PVPS}_k + \text{controls}
\end{equation}

\noindent estimated separately for each emotion $j$ and PVPS axis $k$, with the same controls as the original specification (log followers, log friends, log previous tweets, time fixed effects).

The middle and bottom rows of Figure~\ref{fig:emotion_heterogeneity} show predicted retweet counts as emotions vary from 0 to 10, separately for images classified as favorable to one group or the other. On the Democratic Women vs.\ Republican Men axis, enthusiasm produces a steeper positive slope for Democrat-women-favorable images, while fear and disgust generate stronger engagement for Republican-men-favorable images. Sadness depresses engagement asymmetrically, mainly for Republican-men-favorable content.

The Hispanic Democrat vs.\ Non-Hispanic Republican axis shows the same qualitative structure, though with wider confidence intervals due to the smaller effective sample size. Across additional intersectional axes (Appendix~\ref{sec:app_profiles}), emotional effects consistently depend on which group the image is perceived to favor. Enthusiasm mobilizes aligned audiences, while fear and disgust amplify engagement with threatening outgroup content, patterns obscured by averaged emotion scores alone.

\subsection*{Won, Steinert-Threlkeld, and Joo (2017)}

Won, Steinert-Threlkeld, and Joo (2017) release the UCLA Protest Image Dataset (40,764 images, 11,659 protest) annotated for binary protest presence, ten visual attributes, and continuous perceived violence via Bradley-Terry pairwise comparisons. Their Table 4 (of the original article) reports that fire ($r = +.59$) and law enforcement ($r = +.37$) predict higher perceived violence, while signs ($r = -.49$) predict lower violence. These correlations are treated as properties of the images themselves. The annotators' demographics are not recorded.

\begin{figure}[htbp]
\vspace{1cm}
    \centering
    \includegraphics[width=1.1\textwidth]{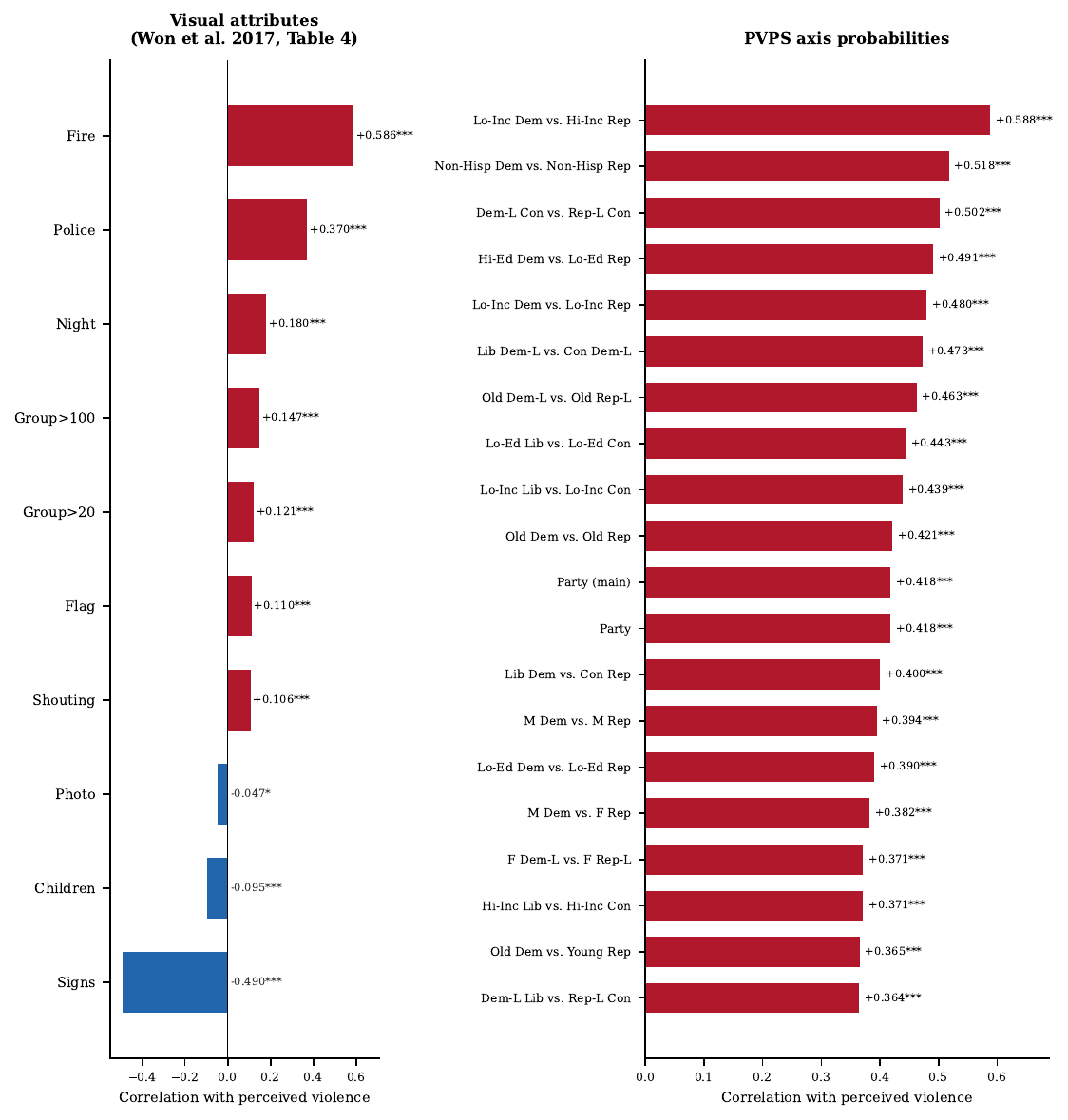}
    \caption{\footnotesize Correlates of perceived violence in protest images ($n = 2{,}343$, UCLA Protest Image Dataset test set). Left panel replicates Won et al.\ (2017) Table 4 showing correlations between visual attributes and perceived violence. Right panel shows correlations between PVPS axis probabilities and the same violence scores. All PVPS axes correlate positively with violence, meaning images perceived as more violent are the same images the PVPS classifier predicts as more favorably evaluated by Republican, conservative, and Republican-leaning respondents. $^{***}p < .001$, $^{**}p < .01$, $^{*}p < .05$.}
    \label{fig:won_table4}
    \vspace{1cm}
\end{figure}

I run 2,343 held-out protest test images (none in the PVPS training data) through the PVPS classifier and correlate each axis probability with perceived violence. Figure~\ref{fig:won_table4} shows the results side by side. The left panel replicates Won et al.'s findings on the test set. The right panel shows that every PVPS axis correlates positively with violence, with the strongest effects on cross-party axes (Low-income Dem.\ vs.\ High-income Rep.\ $r = +.588$, Non-Hispanic Dem.\ vs.\ Non-Hispanic Rep.\ $r = +.518$, Dem.-leaning Con.\ vs.\ Rep.-leaning Con.\ $r = +.502$, all $p < .001$). The primary party axis reaches $r = +.418$. The weakest effect appears on the Dem.-leaning Lib.\ vs.\ Rep.-leaning Con.\ axis ($r = +.364$), but all axes are significant at $p < .001$.
Images that the PVPS classifier predicts as more favorably evaluated by Republicans, conservatives, and Republican-leaning respondents are also the images that Won et al.'s annotators rate as more violent, while images with protest signs and peaceful marches are predicted as more favorably evaluated by Democrats and liberals.

This suggests that perceived violence in protest imagery is not politically neutral but follows the same partisan fault lines that structure political evaluation. Features such as fire and police presence characterize images aligned with conservative audiences, whereas protest signs characterize images aligned with liberal audiences. Because the MTurk annotator pool used by Won et al.\ did not record political demographics, the dataset cannot test whether Democrats and Republicans assign different violence ratings to the same images, a question that a perspectivist annotation design could directly address. Importantly, these results do not invalidate their findings but add an additional layer of structure by showing that visual attributes, audience identity, and evaluative outcomes jointly shape perceived violence.

\section*{Discussion and Limitations}

Political science has long argued that evaluative disagreement is structured by identity and social location (\cite{campbell_american_1960, kinder_us_2009, mason_uncivil_2018}). Computational social science has extended this tradition by applying computer vision and NLP tools to classify political content at scale (\cite{grimmer_textasdata_2022, won2017protest, joo_image_2018}), but these methods were primarily designed for tasks with a single correct label and therefore collapse annotator disagreement by construction (\cite{uma_learning_2021, cabitza_perspectivist_2023}). As a result, they are well suited for prediction tasks but less suited for studying the structure of disagreement itself.

PVPS addresses this gap by turning evaluative disagreement into the object of prediction. It embeds perspectivist assumptions from political science directly into a computational model, which  makes structured disagreement both observable and predictable from image content. However, several design choices constrain what this approach can recover, and these constraints point to clear directions for future work:

\textit{Soft labels as an alternative architecture:}
The PVPS classifier trains a separate binary model for each social axis. An alternative would be to train a single model on the full distribution of annotator responses, known as soft-label training, where the model receives the proportion of annotators who chose each option rather than a single collapsed answer. Soft-label approaches have shown advantages in subjective NLP tasks \citep{fornaciari_beyond_2021, uma_learning_2021, peterson_human_2019, rodrigues_deep_2018}. I chose per-group classifiers because a soft-label distribution captures \textit{that} annotators disagree but not \textit{who} disagrees with whom. This means that an image where Democrats and Republicans split 60/40 in opposite directions produces the same bimodal shape as one where young and old split identically. Per-group classifiers keep the social identity of the disagreement visible \citep{mostafazadeh_davani_dealing_2022}. A hybrid architecture combining distributional training with group-specific heads is a natural next step.

\textit{Dataset scale:}
The primary dataset contains 1,264 images and 82,000 observations. The classifier operates on frozen CLIP and DINOv2 representations pretrained on hundreds of millions of images, so the classification head only learns which feature combinations predict group disagreement. Transfer learning at this scale has proven effective in political science \citep{radford_clip_2021, kornblith_transfer_2019, joo_image_2018, xi_ideology_2020}, and U.S.\ partisan effect sizes are large ($d \approx 1.0$--$1.7$; \cite{iyengar_fear_2015}), meaning the groups are far apart to begin with. The uniform within-party null results confirm the classifier is not picking up noise. More images would sharpen the analysis, but the core partisan signal is strong enough at this scale.

\textit{Other possible social axes:}
Religion \citep{layman_cultural_1997}, geography \citep{rodden_why_2019}, media consumption \citep{prior_post_2007}, and moral foundations \citep{graham_liberals_2009} all structure political attitudes in ways that should matter for image evaluation. Some are partially absorbed by included axes (religiosity correlates with partisanship; geography with education and income). The selection I made follows theory-driven variable choice, but the framework is modular. So any axis can be added given annotator-level data, and whether it matters empirically is itself a finding the classifier produces.

\textit{Higher-order intersections:}
The classifier models pairwise intersections but not three-way (or x-way) combinations. Intersectionality theory motivates attention to joint identity categories \citep{crenshaw_mapping_1991, mccall_complexity_2005}, but detecting a two-way interaction already requires roughly sixteen times the sample size of a main effect \citep{gelman_interaction_2018, mcclelland_detecting_1993}. In practice, two-way intersections capture the most consequential disparities \citep{buolamwini_gender_2018, kearns_fairness_2018}. Larger datasets would make selected three-way intersections feasible, and the pipeline accommodates them without architectural changes.

\textit{Generalizability:}
The U.S.\ is an affective-polarization outlier among OECD democracies \citep{boxell_cross-country_2022}, its two-party system and media environment operate under dynamics that do not transfer straightforwardly to multiparty settings \citep{dalton_quantity_2008, hallin_comparing_2004}, and cross-cultural research in cognitive psychology suggests that even basic visual perception is shaped by cultural context \citep{nisbett_influence_2005, henrich_weirdest_2010}. The principle that stable social cleavages produce structured disagreement should hold across contexts \citep{lipset_cleavage_1967, kriesi_west_2008}, but the images that activate them will differ, since topics like Black Lives Matter are specifically American.

Who will use PVPS? Across social media research \citep{barbera_birds_2015, brady_etal2017, guess_etal2023, jia_etal2025}, PVPS turns image flows into group-conditioned evaluative profiles, making it possible to test whether amplification favors content that increases consensus within groups or divergence across them. In studies of visual framing by news outlets \citep{peng_same_2018, gasparyan_media_2025, geise_visual_2025}, the unit of analysis shifts from what is shown to how different audience segments respond. This allows outlet strategies to be evaluated in terms of anticipated audience polarization. Work on visual misinformation \citep{yang_davis_hindman2023, hameleers2024} can rank images by expected disagreement across groups and identify those most likely to generate partisan split rather than generic false belief. In experimental research \citep{hiaeshutter_rice_cued_2023}, the same mechanism removes the need for iterative pretesting by scoring candidate stimuli directly in terms of expected audience structure.

More generally, PVPS replaces post hoc stimulus selection with a pre-measured mapping of expected evaluative disagreement across social groups. This makes it possible to design visual experiments from the structure of audience disagreement itself.

\newpage

\printbibliography

\newpage
\appendix

\section*{Appendix}

\section{Full Classification Results}
\label{sec:app_full_results}

This appendix presents the complete classification results for all axes tested and explains the accuracy threshold used to determine which axes produce reliable classifiers. Political and cross-political axes use the main PVPS classifier (Figure~\ref{fig:pipeline}). Sociodemographic axes (within-party, within-ideology, within-thermometer, and standalone social) use the extended pipeline (Figure~\ref{fig:stack_v2}).

\subsection{What counts as a working classifier?}
\label{sec:app_threshold}

A classifier is useful when it predicts the correct label substantially more often than a coin flip. For a binary task, chance accuracy is 50\%. I set the working threshold at 65\%, fifteen percentage points above chance, for three reasons.

First, the prediction target differs from standard image classification. The classifier does not identify objects in an image; it predicts which of two political or social location groups rated the image more positively. Each image in the main survey was rated by approximately 65 respondents on average, producing unusually rich ground-truth labels. Most computational work on political images relies on a single annotator or a small crowd panel per item \citep{won2017protest, casas_images_2019}; the present study aggregates dozens of ratings per image, which yields more precise estimates of group-level evaluative differences. Comparable studies that predict political orientation from visual data report accuracies in the 70--75\% range even with far larger training sets. \citet{kosinski_facial_2021} achieve 72\% from over one million facial photographs, and \citet{won2017protest} report AUC $\sim$.97 for protest detection from 40,764 images, a less subjective target than evaluative divergence. The present study works with 1,264 primary images and a more abstract prediction target than object recognition, making 65\% a conservative threshold. \textit{This threshold applies to the minimum per-class accuracy, the accuracy on whichever class is harder to predict. Overall accuracy on passing axes is higher in every case.}

Second, the training set is compact by machine learning standards. Each of the 100 training runs (10 seeds $\times$ 10 ensemble members) uses roughly 1,075 images for training and 189 for testing. Intersectional axes (for example, Female Democrats versus Male Republicans) further reduce the usable sample to 200--400 images. That the classifier achieves 65--83\% accuracy in this regime, without pretraining on millions of labeled examples, suggests that the evaluative signal in political images is strong enough to be captured even from limited data.

Third, the classifier is a measurement tool and apparently not the only decision system. When it is later applied to label new, unseen images, any systematic errors can be corrected statistically using the approach proposed by \citet{egami_using_2024}. What matters is that accuracy is reliably above chance, stable across the 10 runs, and not achieved by always guessing the larger class.\footnote{A classifier that predicts ``consensus'' for every image in a dataset where 60\% of images are consensus would reach 60\% accuracy but would be useless. Per-class accuracy, reported separately for each group, guards against this.}

\subsection{How to read Table~\ref{tab:app_full}}
\label{sec:app_howtoread}

Each row is one classification axis. The columns report the following.

\begin{itemize}
\item \textbf{Grp~A, Grp~B} give the classifier's accuracy on each of the two groups separately. For example, in the row ``Female Dem vs Male Rep,'' Grp~A is the share of Female-Democrat-preferred images correctly identified, and Grp~B is the share of Male-Republican-preferred images correctly identified. If either number is near 50\%, the classifier cannot reliably distinguish that group's images from the other's.
\end{itemize}

{\small
\begin{longtable}{lcc@{\hskip 8pt}c}
\caption{Classification results for all axes. Grp~A and Grp~B give per-class accuracy (\%) for the first and second group named in each axis. $\star$\,=\,both groups above 75\%; $\bullet$\,=\,both above 65\%; $\circ$\,=\,min-class $\geq$ 60\% (Stack~V2 axes only). All values are medians across 10 seeds. DemL\,=\,Dem-leaning (feeling thermometer); RepL\,=\,Rep-leaning. Political and cross-political axes use the main PVPS classifier. Sociodemographic axes (marked Stack~V2) use the extended pipeline.}
\label{tab:app_full} \\
\toprule
\textbf{Axis} & \textbf{Grp A (\%)} & \textbf{Grp B (\%)} & \\
\midrule
\endfirsthead
\multicolumn{4}{l}{\small\textit{Table~\ref{tab:app_full} continued}} \\
\toprule
\textbf{Axis} & \textbf{Grp A (\%)} & \textbf{Grp B (\%)} & \\
\midrule
\endhead
\midrule
\multicolumn{4}{r}{\small\textit{Continued on next page}} \\
\endfoot
\bottomrule
\endlastfoot
\multicolumn{4}{l}{\textit{Primary political axes}} \\
\quad IDEOLOGY                                & 81.6 & 78.9 & $\star$ \\
\quad THERMOMETER                             & 79.8 & 78.5 & $\star$ \\
\quad PARTY                                   & 68.7 & 69.6 & $\bullet$ \\
\midrule
\multicolumn{4}{l}{\textit{Social axes, no partisan conditioning (Stack~V2)}} \\
\quad AGE                                     & 70.5 & 68.9 & $\bullet$ \\
\quad GENDER                                  & 67.4 & 62.0 & $\circ$ \\
\quad EDUCATION                               & 65.3 & 61.0 & $\circ$ \\
\quad HISPANIC                                & 63.8 & 57.5 &  \\
\quad INCOME                                  & 63.3 & 53.5 &  \\
\midrule
\multicolumn{4}{l}{\textit{Cross-party}} \\
\quad Female Dem vs Male Rep                  & 79.4 & 75.5 & $\star$ \\
\quad HighInc Dem vs LowInc Rep               & 81.8 & 81.6 & $\star$ \\
\quad Hisp Dem vs NonHisp Rep                 & 78.1 & 83.8 & $\star$ \\
\quad LowInc Dem vs HighInc Rep               & 83.6 & 79.3 & $\star$ \\
\quad Young Dem vs Old Rep                    & 79.5 & 78.9 & $\star$ \\
\quad Liberal Dem vs Conserv Rep              & 80.9 & 74.7 & $\bullet$ \\
\quad Strong Dem vs Strong Rep               & 80.6 & 74.1 & $\bullet$ \\
\quad HighEdu Dem vs LowEdu Rep              & 82.6 & 73.5 & $\bullet$ \\
\quad Old Dem vs Young Rep                   & 82.5 & 73.1 & $\bullet$ \\
\quad Male Dem vs Female Rep                 & 83.2 & 70.5 & $\bullet$ \\
\quad LowEdu Dem vs HighEdu Rep              & 78.3 & 67.7 & $\bullet$ \\
\quad NonHisp Dem vs Hisp Rep                & 73.7 & 60.0 & \\
\quad Moderate Dem vs Moderate Rep            & 57.1 & 54.5 & \\
\quad WeakDem vs WeakRep                     & 54.0 & 63.5 & \\
\midrule
\multicolumn{4}{l}{\textit{Cross-ideology}} \\
\quad HighEdu Lib vs LowEdu Con              & 86.5 & 80.4 & $\star$ \\
\quad Male Lib vs Female Con                 & 82.9 & 75.0 & $\star$ \\
\quad Old Lib vs Young Con                   & 77.4 & 74.5 & $\bullet$ \\
\quad DemLean Lib vs RepLean Con             & 80.7 & 74.6 & $\bullet$ \\
\quad LowInc Lib vs HighInc Con              & 78.9 & 71.7 & $\bullet$ \\
\quad Female Lib vs Male Con                 & 83.5 & 72.0 & $\bullet$ \\
\quad LowEdu Lib vs HighEdu Con              & 75.0 & 76.8 & $\star$ \\
\quad Young Lib vs Old Con                   & 76.6 & 77.6 & $\star$ \\
\quad HighInc Lib vs LowInc Con              & 77.3 & 72.7 & $\bullet$ \\
\quad Hisp Lib vs NonHisp Con                & 76.1 & 64.7 & \\
\quad NonHisp Lib vs Hisp Con                & 71.8 & 62.8 & \\
\quad RepLean Lib vs DemLean Con             & 45.5 & 44.4 & \\
\midrule
\multicolumn{4}{l}{\textit{Cross-thermometer}} \\
\quad HighEdu DemL vs LowEdu RepL            & 81.3 & 78.8 & $\star$ \\
\quad LowEdu DemL vs HighEdu RepL            & 78.1 & 76.9 & $\star$ \\
\quad Young DemL vs Old RepL                 & 77.3 & 82.8 & $\star$ \\
\quad Female DemL vs Male RepL               & 80.0 & 75.0 & $\star$ \\
\quad Male DemL vs Female RepL               & 81.7 & 75.0 & $\star$ \\
\quad LowInc DemL vs HighInc RepL            & 79.5 & 75.0 & $\star$ \\
\quad HighInc DemL vs LowInc RepL            & 81.6 & 70.2 & $\bullet$ \\
\quad Hisp DemL vs NonHisp RepL              & 82.3 & 70.4 & $\bullet$ \\
\quad Old DemL vs Young RepL                 & 80.7 & 68.3 & $\bullet$ \\
\quad NonHisp DemL vs Hisp RepL              & 63.9 & 57.1 & \\
\midrule
\multicolumn{4}{l}{\textit{Same-sociodemographic, cross-party}} \\
\quad Female Dem vs Female Rep               & 82.1 & 82.0 & $\star$ \\
\quad HighEdu Dem vs HighEdu Rep             & 80.6 & 77.2 & $\star$ \\
\quad LowEdu Dem vs LowEdu Rep              & 79.2 & 73.2 & $\bullet$ \\
\quad Young Dem vs Young Rep                 & 76.1 & 71.7 & $\bullet$ \\
\quad LowInc Dem vs LowInc Rep              & 77.0 & 71.4 & $\bullet$ \\
\quad NonHisp Dem vs NonHisp Rep             & 84.2 & 71.1 & $\bullet$ \\
\quad HighInc Dem vs HighInc Rep             & 80.9 & 71.6 & $\bullet$ \\
\quad Male Dem vs Male Rep                   & 83.3 & 70.6 & $\bullet$ \\
\quad Old Dem vs Old Rep                     & 81.5 & 69.2 & $\bullet$ \\
\quad Hisp Dem vs Hisp Rep                   & 54.2 & 59.0 & \\
\midrule
\multicolumn{4}{l}{\textit{Same-sociodemographic, cross-ideology}} \\
\quad Old Lib vs Old Con                     & 86.2 & 79.7 & $\star$ \\
\quad HighEdu Lib vs HighEdu Con             & 79.6 & 80.0 & $\star$ \\
\quad Female Lib vs Female Con               & 81.0 & 76.9 & $\star$ \\
\quad NonHisp Lib vs NonHisp Con             & 84.4 & 75.0 & $\star$ \\
\quad Male Lib vs Male Con                   & 83.8 & 76.2 & $\star$ \\
\quad HighInc Lib vs HighInc Con             & 76.3 & 73.0 & $\bullet$ \\
\quad LowEdu Lib vs LowEdu Con              & 76.7 & 69.8 & $\bullet$ \\
\quad LowInc Lib vs LowInc Con              & 72.2 & 72.1 & $\bullet$ \\
\quad Young Lib vs Young Con                 & 70.2 & 56.5 & \\
\quad Hisp Lib vs Hisp Con                   & 60.5 & 64.4 & \\
\midrule
\multicolumn{4}{l}{\textit{Same-sociodemographic, cross-thermometer}} \\
\quad HighEdu DemL vs HighEdu RepL           & 80.8 & 77.6 & $\star$ \\
\quad Female DemL vs Female RepL             & 83.6 & 77.6 & $\star$ \\
\quad Old DemL vs Old RepL                   & 82.1 & 76.8 & $\star$ \\
\quad Male DemL vs Male RepL                 & 81.7 & 74.1 & $\bullet$ \\
\quad LowInc DemL vs LowInc RepL            & 79.7 & 73.9 & $\bullet$ \\
\quad LowEdu DemL vs LowEdu RepL            & 77.8 & 74.0 & $\bullet$ \\
\quad HighInc DemL vs HighInc RepL           & 81.2 & 78.8 & $\star$ \\
\quad NonHisp DemL vs NonHisp RepL           & 77.5 & 71.6 & $\bullet$ \\
\quad Young DemL vs Young RepL               & 73.0 & 69.8 & $\bullet$ \\
\quad Hisp DemL vs Hisp RepL                 & 45.8 & 55.9 & \\
\midrule
\multicolumn{4}{l}{\textit{Within-party (Stack~V2)}} \\
\quad AGE                                    & 70.5 & 68.9 & $\bullet$ \\
\quad Young Rep vs Old Rep                   & 66.7 & 65.4 & $\bullet$ \\
\quad Young Dem vs Old Dem                   & 65.5 & 63.6 & $\circ$ \\
\quad GENDER                                 & 67.4 & 62.0 & $\circ$ \\
\quad EDUCATION                              & 65.3 & 61.0 & $\circ$ \\
\quad HISPANIC                               & 63.8 & 57.5 &  \\
\quad Female Rep vs Male Rep                 & 65.9 & 57.3 &  \\
\quad Female Dem vs Male Dem                 & 58.5 & 56.8 &  \\
\quad HighEdu Dem vs LowEdu Dem              & 61.7 & 56.8 &  \\
\quad INCOME                                 & 63.3 & 53.5 &  \\
\quad Hisp Dem vs NonHisp Dem                & 60.0 & 53.0 &  \\
\quad Hisp Rep vs NonHisp Rep                & 60.0 & 52.4 &  \\
\quad HighEdu Rep vs LowEdu Rep              & 55.7 & 51.1 &  \\
\midrule
\multicolumn{4}{l}{\textit{Within-ideology (Stack~V2)}} \\
\quad DemLean Con vs RepLean Con             & 70.3 & 64.2 & $\circ$ \\
\quad Young Con vs Old Con                   & 69.1 & 63.9 & $\circ$ \\
\quad Lib DemL vs Con DemL                   & 68.7 & 63.0 & $\circ$ \\
\quad Hisp Con vs NonHisp Con                & 62.2 & 56.4 &  \\
\quad Female Con vs Male Con                 & 60.4 & 56.1 &  \\
\quad Young Lib vs Old Lib                   & 56.8 & 53.4 &  \\
\quad HighEdu Lib vs LowEdu Lib              & 58.3 & 52.3 &  \\
\quad Female Lib vs Male Lib                 & 57.3 & 51.8 &  \\
\quad HighInc Lib vs LowInc Lib              & 54.1 & 49.8 &  \\
\quad HighInc Con vs LowInc Con              & 58.0 & 52.9 &  \\
\quad HighEdu Con vs LowEdu Con              & 50.5 & 48.0 &  \\
\quad Hisp Lib vs NonHisp Lib                & 50.0 & 42.1 &  \\
\quad DemLean Lib vs RepLean Lib             & \multicolumn{2}{c}{\textit{collapsed}} &  \\
\midrule
\multicolumn{4}{l}{\textit{Within-thermometer (Stack~V2)}} \\
\quad Young DemL vs Old DemL                 & 63.3 & 59.5 &  \\
\quad Young RepL vs Old RepL                 & 63.0 & 59.2 &  \\
\quad Female RepL vs Male RepL               & 61.2 & 55.9 &  \\
\quad Female DemL vs Male DemL               & 57.3 & 54.5 &  \\
\quad HighEdu DemL vs LowEdu DemL            & 62.1 & 51.1 &  \\
\quad HighEdu RepL vs LowEdu RepL            & 58.1 & 50.7 &  \\
\quad HighInc DemL vs LowInc DemL            & 58.1 & 54.4 &  \\
\quad HighInc RepL vs LowInc RepL            & 57.7 & 51.6 &  \\
\quad Hisp DemL vs NonHisp DemL              & 53.8 & 46.8 &  \\
\quad Hisp RepL vs NonHisp RepL              & 55.2 & 48.7 &  \\
\quad Lib RepL vs Con RepL                   & \multicolumn{2}{c}{\textit{collapsed}} &  \\
\end{longtable}
}

\section{Calibration and Prediction Stability}
\label{sec:app_diagnostics}

Table~\ref{tab:diagnostics} reports standard classification diagnostics for the three primary political axes.

\begin{table}[htbp]
\centering
\small
\caption{Classification diagnostics for primary political axes. All values are medians across 10 seeds. Accuracy, precision, recall, and F1 are reported as macro-averages (unweighted mean across the two classes). AUC is the area under the ROC curve.}
\label{tab:diagnostics}
\vspace{4pt}
\begin{tabular}{lccccc}
\toprule
\textbf{Axis} & \textbf{Accuracy} & \textbf{Precision} & \textbf{Recall} & \textbf{Macro F1} & \textbf{AUC} \\
\midrule
PARTY               & 84.0\% & 75.4\% & 82.5\% & 77.4\% & 0.874 \\
IDEOLOGY            & 87.1\% & 81.9\% & 83.3\% & 82.5\% & 0.906 \\
THERMOMETER         & 88.7\% & 85.2\% & 86.4\% & 86.4\% & 0.915 \\
\bottomrule
\end{tabular}
\end{table}

The spread (interquartile range) across 10 seeds is below 4 percentage points on most passing axes, indicating that results are not driven by a lucky train-test split. The primary political axes show spread between 2.4pp (ideology) and 3.8pp (thermometer). Within-group axes that cluster near chance show equally low spread, confirming that these null results are stable, not noisy.

The ensemble's predicted probabilities for the PARTY axis are bimodal. 69\% of test-image predictions fall in the tails (below 0.2 or above 0.9), with 53\% of images receiving near-zero probability of class~1 and 14\% receiving near-certainty. The classifier is confidently correct on most images, with roughly 20\% in the uncertain middle range (0.2--0.8).

\section{Feature Importance and Ablation}
\label{sec:app_ablation}

Table~\ref{tab:ablation} reports the ablation study on the PARTY axis (primary data only, 10 seeds per configuration), removing one feature block at a time.

\begin{table}[htbp]
\centering
\small
\caption{Feature ablation on the PARTY axis (primary data only). Each row removes one feature block; other blocks remain. Drop = change from full-model baseline.}
\label{tab:ablation}
\vspace{4pt}
\begin{tabular}{lcc}
\toprule
\textbf{Configuration} & \textbf{Median min-acc} & \textbf{Drop} \\
\midrule
Full model & 73.2\% & -- \\
$-$ CLIP & 70.3\% & $-$2.8pp \\
$-$ DINOv2 & 71.3\% & $-$1.9pp \\
$-$ TF-IDF + Semantic & 72.9\% & $-$0.2pp \\
$-$ Concepts & 70.2\% & $-$3.0pp \\
CLIP only & 71.8\% & $-$1.4pp \\
DINOv2 only & 69.8\% & $-$3.3pp \\
\bottomrule
\end{tabular}
\caption{Feature ablation on the PARTY axis (primary data only, single-ensemble setup with 10 seeds per row). The full-model baseline of 73.2\% reflects this single-ensemble configuration; the corresponding median in the 10-ensemble × 10-seed configuration used elsewhere in the paper is 68.7\% (Table~\ref{tab:app_full}). Drop = change from the full-model baseline within this configuration.}
\end{table}

Removing concept features (binary indicators of political objects such as protest signs, police, flags, and politicians) produces the largest drop ($-$3.0pp), followed by CLIP ($-$2.8pp) and DINOv2 ($-$1.9pp). Text-based features contribute the least ($-$0.2pp when TF-IDF captions and semantic word indicators are removed together), suggesting that the partisan signal on this axis is carried by visual content. CLIP alone (71.8\%) and DINOv2 alone (69.8\%) each approach the full-model baseline without matching it, which means the two pretrained encoders capture complementary aspects of the image. The concept features add information that neither encoder represents on its own, likely because these features were constructed for political image content and encode objects (uniforms, banners, crowd formations) that general-purpose vision models treat as incidental.

The feature vector comprises six blocks totaling 1,666 dimensions. CLIP (512) and DINOv2 (768) are pure visual embeddings extracted from frozen pretrained models. TF-IDF caption features (300) encode word frequencies in Gemini-generated image descriptions. Semantic word indicators (50) flag the presence of politically predictive terms identified during training. Concept presence features (24) capture high-level scene attributes (see Section~\ref{sec:app_concepts}), and political topic features (12) indicate policy-domain relevance.

\section{Source Prediction Diagnostic}
\label{sec:app_source}

The pipeline includes a pre-training diagnostic that tests whether a simple logistic regression on CLIP and DINOv2 embeddings can distinguish primary-data images from supplementary \citet{williams2026conservatives} images. If source-classification accuracy exceeds 85\%, self-training is disabled automatically.

The diagnostic returns 97.3\% accuracy (5 splits, range 96.9--97.4\%). The two image pools differ substantially in resolution, framing, and photographic style. For image-level classifiers (the main pipeline of Figure~\ref{fig:pipeline} and Path~A of the extension), self-training was disabled and the supplementary dataset was excluded from training, since image-level classification could otherwise learn dataset-style differences as spurious signal. Path~B of the extension trains on individual (image, respondent) pairs and is fitted on the combined respondent-level pool, which is statistically appropriate because dataset-style differences cannot enter as a per-respondent rule.

\section{Per-Topic Sociodemographic Classification}
\label{sec:app_pertopic}

When the training data are split by policy topic, sociodemographic axes reveal stable directional patterns that cancel in the aggregate. Table~\ref{tab:app_pertopic} reports per-topic results for each standalone sociodemographic axis. Where 75\%+ of images within a topic have the same directional label, the finding is reported as a dominant direction. Where both classes are represented, a classifier is trained on balanced holdout splits (equal class sizes in test, minimum five per class, ten seeds) and median minimum-class accuracy is reported.

\begin{table}[htbp]
\centering
\small
\caption{Per-topic sociodemographic classification. ``Dom.'' = dominant direction ($\geq$75\% of images have same label). Accuracy = median minimum-class accuracy, 10 seeds, balanced test.}
\label{tab:app_pertopic}
\vspace{4pt}
\begin{tabular}{lcccc}
\toprule
\textbf{Axis} & \textbf{Immigration} & \textbf{Guns} & \textbf{LGBTQ+} & \textbf{Jan.~6} \\
\midrule
GENDER    & 52.3\%               & Dom.\ (F, 76\%)   & Dom.\ (M, 78\%)    & 22.2\% \\
HISPANIC  & Dom.\ (Hisp, 76\%)   & 0.0\%             & Dom.\ (Hisp, 82\%) & 25.0\% \\
AGE       & 45.2\%               & 14.3\%            & 40.9\%             & Dom.\ (Young, 76\%) \\
EDUCATION & 51.5\%               & 25.0\%            & 33.3\%             & Dom.\ (LowEdu, 82\%) \\
INCOME    & 43.4\%               & 46.2\%            & 33.3\%             & 41.7\% \\
\bottomrule
\end{tabular}
\end{table}

Where the image-level classifier is tested on balanced holdout sets within a single topic, it performs at or below chance. Gender on immigration reaches 52.3\% (966 images, balanced test of 65 per class). Education on immigration reaches 51.5\% (906 images). Age on immigration reaches 45.2\%. Income falls below chance on every topic. The dominant direction findings confirm that sociodemographic evaluative divergence is real and topic-structured, but the image-level classifier cannot predict from visual content alone which way the demographic gap will go on a specific image.

\section{Encoder Architecture}
\label{sec:app_encoders}

Both image encoders (CLIP ViT-B/32 and DINOv2 ViT-B/14) are used as frozen feature extractors. No fine-tuning is performed on any encoder. Images are passed through each pretrained model to obtain fixed-length embedding vectors (512 dimensions from CLIP, 768 from DINOv2), which are L2-normalized and concatenated with the four text and concept feature blocks before being fed to the classification head.

The classification head is a two-block residual network with GELU activations, layer normalization, and dropout. Input is projected to a hidden dimension (128 or 256, selected by cross-validation), passed through two residual blocks, and mapped to class logits. Ten such networks are trained per seed with different random initializations, and their softmax outputs are averaged to produce the ensemble prediction.

\section{Concept Feature Definitions}
\label{sec:app_concepts}

The 24 concept features are binary indicators of political objects and scene attributes. Each feature is set to 1 if the corresponding object or attribute is detected in the Gemini-generated image description, and 0 otherwise. The concepts were drawn from the political image literature \citep{casas_images_2019, won2017protest, torres2022learning} and refined during development to retain those that contributed to classification accuracy.

The 24 concepts are listed below.

\vspace{-1cm}
\begin{singlespace}
\begin{enumerate}\itemsep0pt\parskip0pt
\item American flag
\item Confederate flag
\item Pride rainbow flag
\item Trump sign or hat
\item BLM sign
\item Police
\item Military
\item Protesters
\item Crowd of people
\item Armed people
\item Politicians
\item Guns or weapons
\item Signs or banners
\item Capitol building
\item Church
\item Border wall or fence
\item Violence or conflict
\item Peaceful gathering
\item Voting or election
\item Immigration or migrants
\item Same-sex couple
\item Angry people
\item Happy people
\item Sad or mourning
\end{enumerate}
\end{singlespace}

Each concept is detected by keyword matching against the structured Gemini description (dense description, objects and symbols, political context fields). A concept is coded 1 if any of its trigger keywords appear in the combined text and 0 otherwise.

The 12 political topic features indicate whether the image description contains keywords associated with a policy domain. The 12 domains are climate, economy and labor, education, elections, gender and reproductive rights (covering MeToo, women's march, abortion, Roe v.\ Wade, Planned Parenthood, reproductive rights, and Kavanaugh hearings as a single dimension), guns, healthcare, immigration, LGBTQ+, military and foreign policy, race and policing, and Trump administration. These features help the classifier differentiate images from distinct policy contexts. Wave~4 includes January~6 imagery (see Table~\ref{tab:data_comparison}); these images do not have a dedicated topic-feature dimension and instead activate the closest semantically related dimensions (Trump administration, race and policing, or elections) depending on caption content.

\section{Full Evaluative Profiles}
\label{sec:app_profiles}

\begin{figure}[htbp]
\centering
\caption{Full evaluative profiles}
\label{fig:app_prof1}
\includegraphics[width=1.1\textwidth]{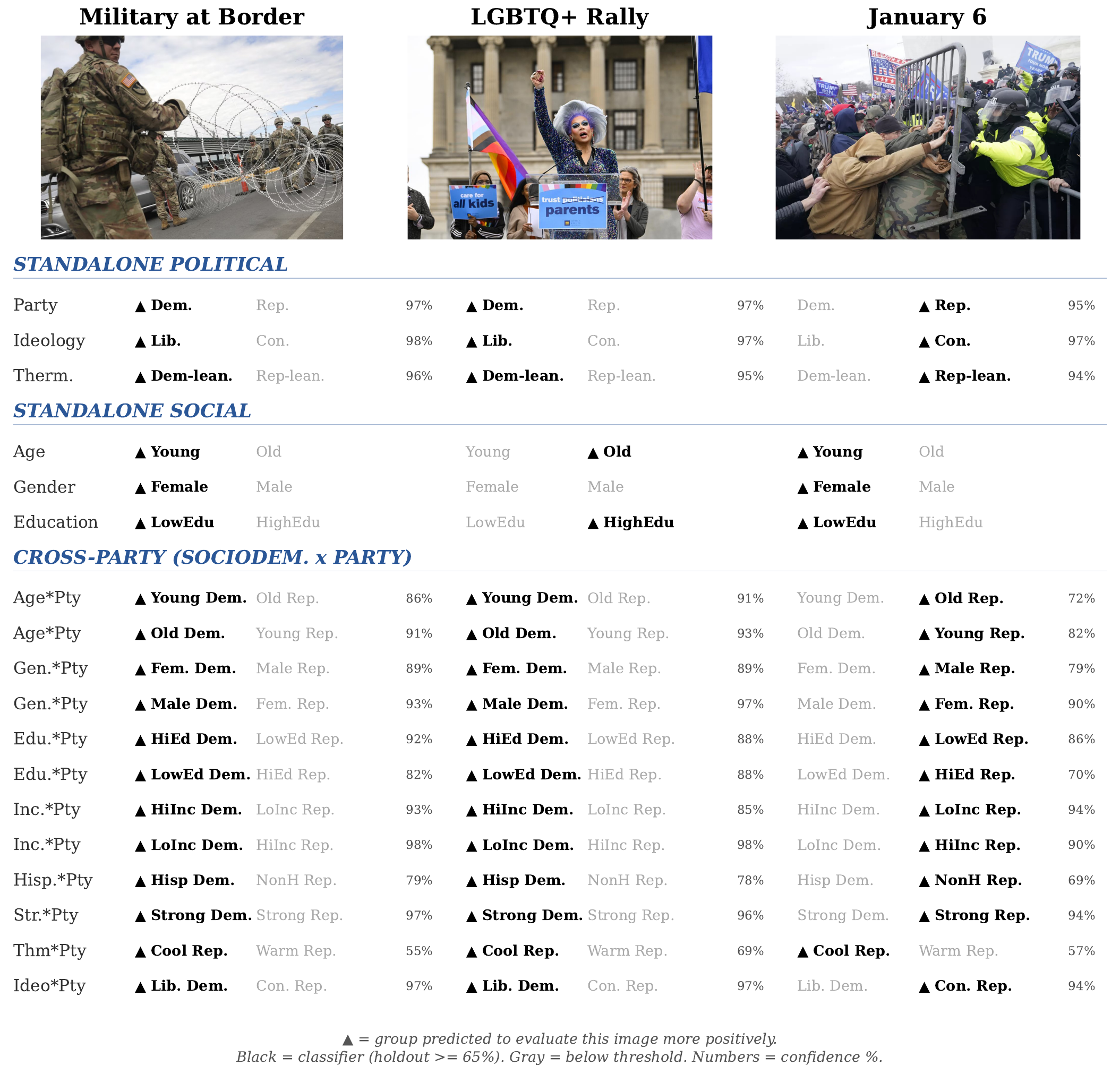}
\end{figure}

Figures~\ref{fig:app_prof1} display the complete evaluative profiles across all axes for three held-out test images (military at border, LGBTQ+ rally, January~6). Black bold triangles mark predictions from the main classifier (holdout accuracy $\geq$ 65\%, applicable to any image). Blue bold triangles mark predictions from the extended sociodemographic pipeline (Stack~V2, median min-class accuracy $\geq$ 60\%). Gray rows indicate axes that do not reach either threshold for the given image.

\begin{figure}[htbp]
\centering
\includegraphics[width=1.1\textwidth]{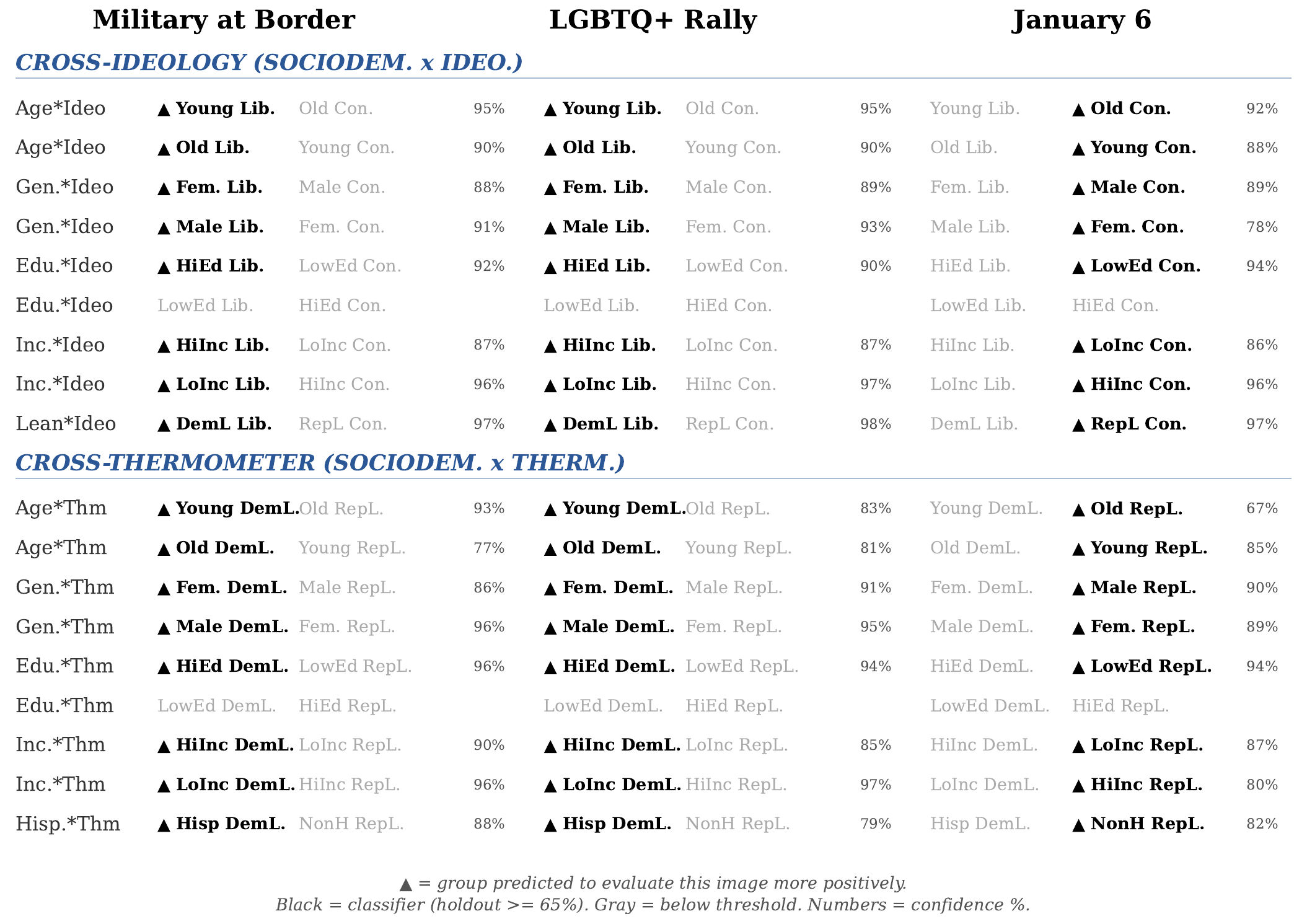}
\label{fig:app_prof2}
\end{figure}

\begin{figure}[htbp]
\centering
\includegraphics[width=1.1\textwidth]{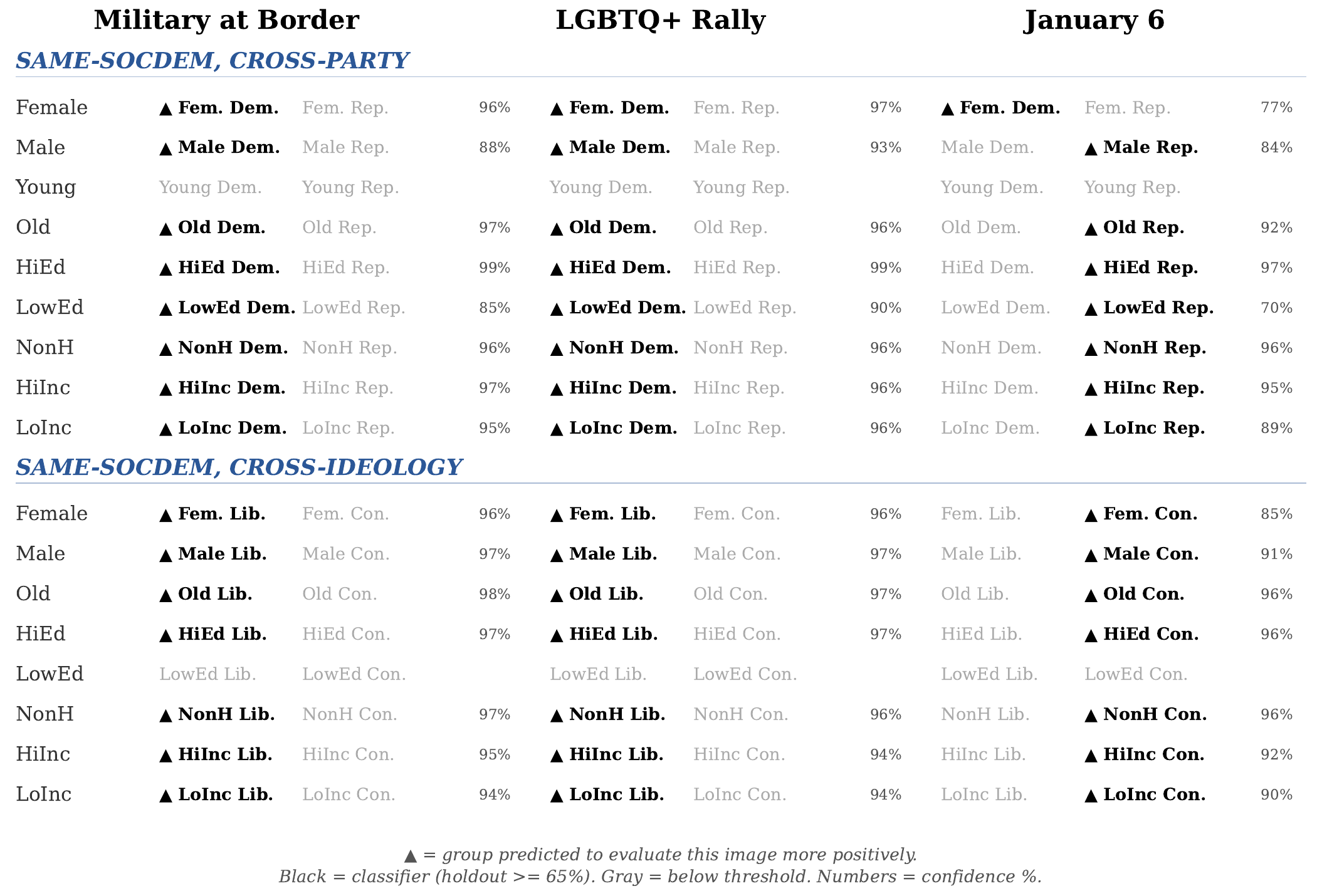}
\label{fig:app_prof3}
\end{figure}

\begin{figure}[htbp]
\centering
\includegraphics[width=1.1\textwidth]{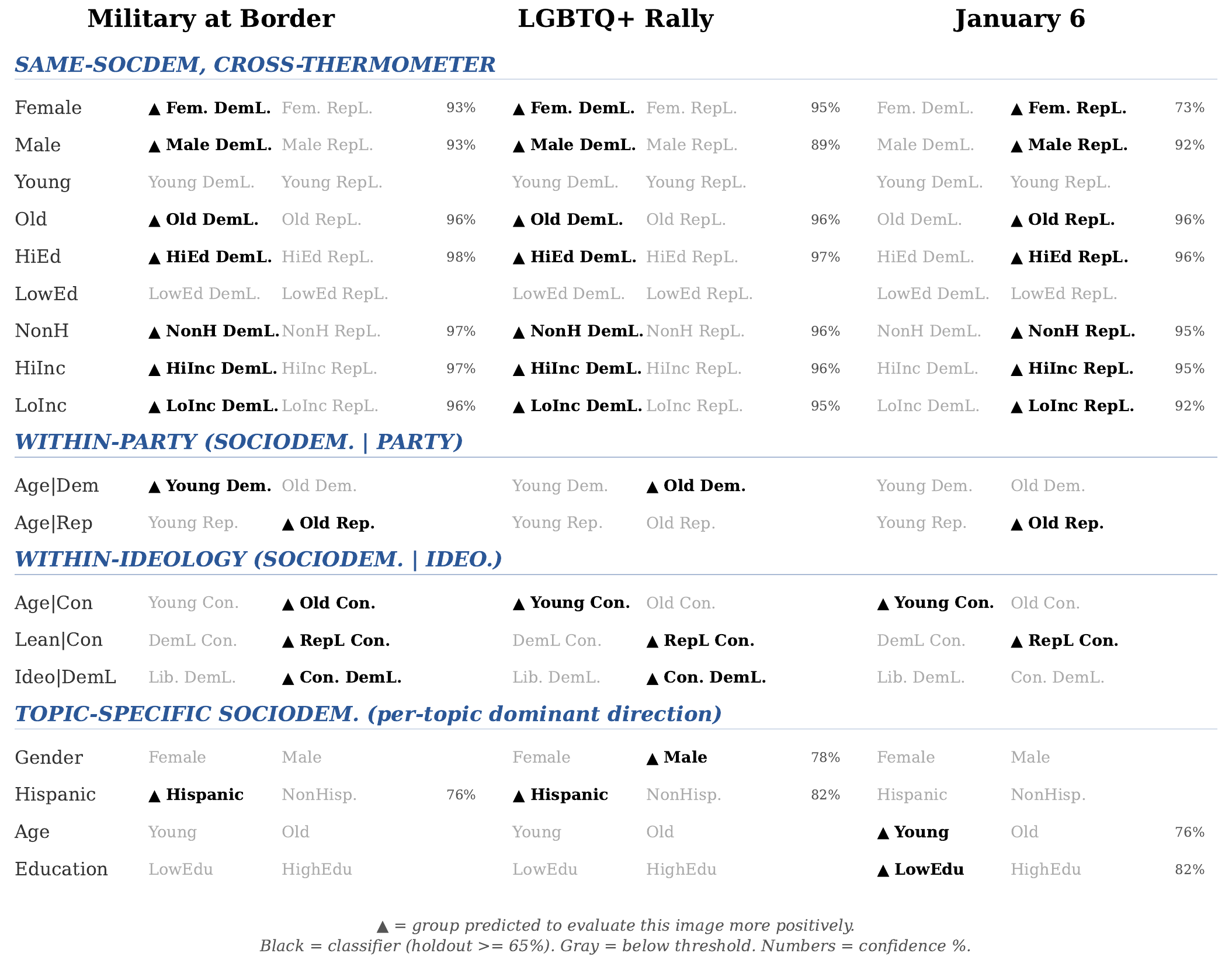}
\label{fig:app_prof4}
\end{figure}

\newpage

\section{Summary of Passing Axes}
\label{sec:app_summary}

\begin{table}[H]
\centering
\small
\caption{Summary of axes by category. Political axes use the main PVPS classifier (Figure~\ref{fig:pipeline}); the 65\% threshold applies. Sociodemographic axes use the extended pipeline (Figure~\ref{fig:stack_v2}); the 60\% threshold applies. Pass = median min-class accuracy above the relevant threshold.}
\label{tab:summary}
\vspace{4pt}
\begin{tabular}{lccc}
\toprule
\textbf{Category} & \textbf{Threshold} & \textbf{Pass / Total} & \textbf{Pass rate} \\
\midrule
Primary political & 65\% & 3 / 3 & 100\% \\
Cross-party & 65\% & 11 / 14 & 79\% \\
Cross-ideology & 65\% & 9 / 12 & 75\% \\
Cross-thermometer & 65\% & 9 / 10 & 90\% \\
Same-socdem, cross-party & 65\% & 9 / 10 & 90\% \\
Same-socdem, cross-ideology & 65\% & 8 / 10 & 80\% \\
Same-socdem, cross-thermometer & 65\% & 9 / 10 & 90\% \\
\midrule
Within-party (Stack~V2) & 60\% & 5 / 13 & 38\% \\
Within-ideology (Stack~V2) & 60\% & 3 / 12 & 25\% \\
Within-thermometer (Stack~V2) & 60\% & 0 / 11 & 0\% \\
Standalone social (Stack~V2) & 60\% & 3 / 5 & 60\% \\
\bottomrule
\end{tabular}
\end{table}

\end{document}